\documentclass[a4paper,fleqn]{cas-sc}

\usepackage[numbers]{natbib} % Numeric citation style (e.g., [1], [2])
\usepackage[nolist]{acronym}

\usepackage{amsmath}
\usepackage{amssymb}

\usepackage{tikz}
\usepackage{xcolor}
\usepackage{graphicx}
\usetikzlibrary{arrows.meta, positioning, calc}
\usepackage{chngcntr}

\usepackage{booktabs}

\definecolor{fwink}{HTML}{1B4F72}
\definecolor{fwmid}{HTML}{4E7EA8}
\definecolor{fwband}{HTML}{EDF3F8}
\definecolor{fwedge}{HTML}{A9C0D6}

\usepackage{xurl}

\newcommand{\frameworkdivider}[2]{%
  \vspace{0.7em}
  \noindent\textbf{#1.} #2\par
  \vspace{0.45em}
}

\newcommand{\fwwhy}{\noindent\textbf{Relevance.} }

\begin{document}
% =========================
% Electrical engineering / protection terms
% =========================

\newacro{pc}[PC]{protection coordination}
\newacro{pr}[PR]{protection relay}
\newacro{fd}[FD]{fault detection}
\newacro{fc}[FC]{fault classification}
\newacro{fli}[FLI]{fault line identification}
\newacro{fl}[FL]{fault localization}

\newacro{emt}[EMT]{electromagnetic transient}
\newacro{rms}[RMS]{root-mean-square}
\newacro{pmu}[PMU]{phasor measurement unit}
\newacro{hil}[HIL]{hardware-in-the-loop}

\newacro{der}[DER]{distributed energy resources}
\newacro{res}[RES]{renewable energy sources}

% =========================
% Machine learning / data terms
% =========================

\newacro{ml}[ML]{machine learning}
\newacro{dl}[DL]{deep learning}
\newacro{ai}[AI]{artificial intelligence}
\newacro{iec}[IEC]{International Electrotechnical Commission}

% Models
\newacro{gb}[GB]{histogram-based gradient boosting}
\newacro{knn}[KNN]{k-nearest neighbors}
\newacro{lg}[LG]{logistic regression}
\newacro{lr}[LR]{linear regression}
\newacro{mlp}[MLP]{multi-layer perceptron}
\newacro{rf}[RF]{random forest}
\newacro{ridge}[Ridge]{ridge regression}

% Optional, if used later
\newacro{hpo}[HPO]{hyperparameter optimization}
\newacro{cv}[CV]{cross-validation}
\newacro{i/o}[I/O]{input/output}

% =========================
% Metrics
% =========================

\newacro{mae}[MAE]{mean absolute error}
\newacro{rmse}[RMSE]{root mean squared error}
\newacro{f1}[F1]{F1 score}
\newacro{rtwo}[$R^2$]{coefficient of determination}

% Optional, if you want explicit metric variants
\newacro{macrof1}[Macro-F1]{macro-averaged F1 score}

% =========================
% Power-system technologies
% =========================

\newacro{ac}[AC]{alternating current}
\newacro{dc}[DC]{direct current}
\newacro{hvdc}[HVDC]{high-voltage direct current}
\newacro{tcsc}[TCSC]{thyristor-controlled series compensation}
\newacro{dfig}[DFIG]{doubly fed induction generator}

% Instrument transformers
\newacro{ct}[CT]{current transformer}
\newacro{vt}[VT]{voltage transformer}

% =========================
% Signal processing and models
% =========================

\newacro{modwt}[MODWT]{maximal overlap discrete wavelet transform}
\newacro{xgboost}[XGBoost]{extreme gradient boosting}
\newacro{cnn}[CNN]{convolutional neural network}
\newacro{cnn1d}[1D-CNN]{one-dimensional convolutional neural network}

% =========================
% Fault-family abbreviations
% =========================

\newacro{slg}[SLG]{single-line-to-ground fault}
\newacro{ll}[LL]{line-to-line fault}
\newacro{llg}[LLG]{double-line-to-ground fault}
\newacro{lll}[LLL]{three-phase fault}
\let\WriteBookmarks\relax
\def\floatpagepagefraction{1}
\def\textpagefraction{.001}
\shorttitle{Framework for Machine Learning in Power System Protection}
\shortauthors{J. Oelhaf et al.}

\title [mode = title]{A Standardized Framework for Machine Learning in Power System Protection}

\author[1]{Julian Oelhaf}[orcid=0009-0008-8204-589X]
\cormark[1] % Corresponding author indication
\ead{julian.oelhaf@fau.de}
\ead[url]{https://lme.tf.fau.de}
\credit{Conceptualization, Methodology, Investigation, Writing -- original draft}

\author[2]{Georg Kordowich}[orcid=0000-0003-2225-7926]
\ead{georg.kordowich@fau.de}
\ead[url]{https://ees.tf.fau.de}
\credit{Conceptualization, Writing - review \& editing}

\author[1]{Paula Andrea Pérez-Toro}[orcid=0000-0002-2727-2116]
\ead{paula.andrea.perez@fau.de}
\credit{Conceptualization, Writing - review \& editing}

\author[3]{Christian Bergler}
\ead{c.bergler@oth-aw.de}
\ead[url]{https://oth-aw.de}
\credit{Writing - review \& editing, Supervision}

\author[2]{Johann Jäger}
\ead{johann.jaeger@fau.de}
\credit{Supervision, Project administration, Funding acquisition}

\author[1]{Andreas Maier}[orcid=0000-0002-9550-5284]
\ead{andreas.maier@fau.de}
\credit{Supervision, Project administration, Funding acquisition}

\author[1]{Siming Bayer}[orcid=0000-0003-2874-4805]
\ead{siming.bayer@fau.de}
\credit{Supervision, Project administration, Funding acquisition, Writing -- review and editing}

\cortext[1]{Corresponding author.}

\affiliation[1]{
  organization={Pattern Recognition Lab, Friedrich-Alexander-Universität Erlangen-Nürnberg},
  addressline={Martensstr. 3},
  postcode={91058},
  city={Erlangen},
  country={Germany}
}

\affiliation[2]{
  organization={Institute of Electrical Energy Systems, Friedrich-Alexander-Universität Erlangen-Nürnberg},
  addressline={Cauerstr. 4},
  postcode={91058},
  city={Erlangen},
  country={Germany}
}

\affiliation[3]{
  organization={Department of Electrical Engineering, Media and Computer Science, Ostbayerische Technische Hochschule Amberg-Weiden},
  addressline={Kaiser-Wilhelm-Ring 23},
  postcode={92224},
  city={Amberg},
  country={Germany}
}

\begin{abstract}
Studies of machine-learning-based power-system protection increasingly report near-perfect scores, yet the meaning of those scores depends strongly on the evaluation setting. Protection task, physical scope, measurements, timing, targets, preprocessing, and validation often vary jointly and remain incompletely specified. This paper proposes a standardization-oriented framework that treats evaluation design as part of the scientific contribution. It defines seven required study dimensions: protection objective, physical system scope, observability and measurements, timing and decision windows, targets and valid samples, training and validation protocol, and evaluation outputs. The framework is instantiated in a bounded case study on the public PROTECT-90 electromagnetic-transient benchmark, comprising 9022 simulated episodes from one 90\,kV double-line topology for onset-conditioned fault classification and fault localization. Under centralized sensing, simulation-metadata-aligned 20\,ms windows, and episode-grouped validation, a multi-layer perceptron (MLP) achieved a five-fold mean macro-averaged F1 score of $0.991 \pm 0.001$ for classification and a localization mean absolute error of $10.20 \pm 0.25\%$ of line length, where the standard deviations describe variation across the episode-grouped folds. Extending the decision horizon to 50\,ms preserved this task-dependent performance asymmetry, while reduced observability approximately doubled the MLP localization error but had little effect on classification. A synchronized two-ended conventional locator outperformed the learning locators under its richer clean information set, and measurement degradation showed that clean predictive performance did not determine robustness. The framework turns evaluation assumptions into explicit and reproducible evidence and provides a basis for more comparable, auditable research evaluation and future certification-oriented assessment of machine-learning protection functions.
\end{abstract}


\maketitle

\section{Introduction}
\label{sec:intro}

Power system protection is a safety-critical function: protective relays must decide correctly within milliseconds, and in practice they are trusted only after standardized type-testing and certification against explicitly specified behavior. Two simultaneous shifts now strain this basis of trust -- the changing physics of the grid, and a change in the methods proposed to protect it. The transition toward decentralized power systems, driven by growing renewable energy sources and distributed energy resources, is reshaping grid operation. Rising shares of inverter-based generation and the adoption of hybrid \ac{ac}--\ac{dc} architectures~\cite{protection_and_automation_b5_protection_2015} expand the range of operating and fault scenarios encountered in practice~\cite{vde_zellulare_2015}. Meshed topologies, multi-terminal configurations, dynamic redispatch, and varying grid-connected or islanded operating modes increase protection-relevant uncertainty, while inverter-based resources contribute fault currents that differ from those of synchronous machines and challenge established protection functions~\cite{schindler_secure_2020,chen_electrical_2005,haddadi_impact_2021,quispe_transmission_2022}. Consequently, conventional protection schemes based on deterministic logic, fixed thresholds, and static system models are increasingly stressed under variable operating conditions~\cite{blackburn_protective_2014}. This has motivated growing interest in data-driven and \ac{ml}-based approaches that can exploit nonlinear structure in protection-relevant measurements. Yet, unlike the conventional functions they aim to augment or replace, these \ac{ml}-based approaches enter the field with no established basis for testing, auditing, or certifying them -- precisely the discipline that makes conventional protection trustworthy.

In protection engineering, performance claims matter only insofar as they can be verified under defined operating, sensing, timing, and failure conditions. A protection function is not trusted because it performs well in one isolated experiment, but because its behavior can be inspected, repeated, and tested against explicit assumptions.
In conventional practice, this trust is established through standardized type-testing and certification: protection equipment is evaluated against common requirements such as \ac{iec}~60255~\cite{international_electrotechnical_commission_measuring_2022} using dedicated relay-test tooling and third-party conformity assessment before deployment. Safety-critical domains are now extending analogous assurance and certification regimes to machine learning -- aviation learning-assurance guidance~\cite{european_union_aviation_safety_agency_artificial_2024}, cross-sector \ac{ai} risk management~\cite{tabassi_artificial_2023}, auditable \ac{ai}-management-system standards~\cite{isoiec_information_2023,isoiec_framework_2022}, and safety-case standards for autonomous products~\cite{ul_standards__engagement_standard_2023} -- alongside a growing literature on assuring the machine-learning lifecycle~\cite{ashmore_assuring_2022}. No widely adopted protection-specific framework currently integrates the requirements needed to evaluate, report, and audit such \ac{ml}-based systems.
This requirement is not yet reflected in much of the \ac{ml}-based protection literature. Public datasets, released code, and high reported scores are useful first steps, but they do not by themselves create deployment-relevant evidence. A model result becomes interpretable only when the protection task, information available at decision time, data-processing pipeline, and validation protocol are specified together. Otherwise, reported performance remains inseparable from hidden study assumptions, and individual papers cannot accumulate into reliable engineering knowledge.

Despite this growing research activity, reported results in \ac{ml}-based protection remain difficult to interpret and compare. Existing studies often differ simultaneously in physical system scope, sensing assumptions, temporal representation, target construction, and validation protocol. These choices are not merely implementation details; they define the effective inference problem by determining which information is available, which timing constraints apply, and what the model is asked to predict. Reported performance therefore reflects not only algorithmic design, but also differences in task formulation and information availability. When these factors vary jointly with model choice, empirical comparisons conflate task difficulty, data-generation choices, and learning effects. A reported accuracy of 99\% in one study may correspond to a comparatively simple task on a radial feeder, whereas 90\% in another may reflect a more demanding setting. Without a common evaluation structure, the community cannot distinguish algorithmic progress from differences in the underlying benchmark.

This paper places evaluation design at the center of the methodological contribution rather than as a neutral experimental backdrop. The central question is not only whether \ac{ml}-based protection can work, but under which physical, observational, and temporal assumptions reported success or failure should be interpreted. This question is especially important because most studies rely on simulated datasets whose physical fidelity, scenario coverage, and documentation vary substantially across works. Without a framework that fixes and reports the core study assumptions, these factors cannot be separated consistently across studies. This work therefore does not propose a new protection method, nor a ranking of models on one benchmark. Its contribution is a standardized evaluation and reporting framework that makes the effective inference problem explicit before performance is interpreted~\cite{mederer_verification_2025}.

\subsection{Limitations of Current Evaluation Practice and Related Work}

A substantial body of literature has applied \ac{ml}-based methods to protection tasks such as \ac{fd}, \ac{fli}, \ac{fc}, and \ac{fl}. For transmission-line protection, wavelet-derived features and artificial neural networks have been used for ultrafast fault detection~\cite{abdullah_ultrafast_2018}, while artificial neural networks and convolutional neural networks have been compared for fault-type identification from voltage and current measurements~\cite{kumar_deep_2022}. Controlled comparative work has also evaluated multiple learning methods for fault detection and line identification under a shared experimental setting~\cite{oelhaf_systematic_2025}. A later controlled comparison on the benchmark used in the present case study examined fault classification and localization under shared sensing, timing, and validation assumptions~\cite{oelhaf_controlled_2026}.

Combined classification and localization have been studied for three-terminal transmission circuits~\cite{livani_fault_2013}, while support-vector-machine models embedded in distribution relays have been used to classify faults, estimate their region, and support switching decisions on a simulated feeder~\cite{jones_machine_2021}. Other studies have addressed joint fault detection, classification, and localization using optimized tree-based and neuro-fuzzy methods~\cite{najafzadeh_fault_2024}, as well as machine-learning-based electrical fault detection and localization in broader grid settings~\cite{vivek_electrical_2024}.

Application-specific protection schemes have further included a \ac{modwt}--\ac{xgboost} pipeline for fault detection and classification across changing microgrid topologies and operating modes~\cite{patnaik_modwt-xgboost_2021}, and a hybrid protection scheme based on deep reinforcement learning~\cite{kordowich_hybrid_2022}. Graph-neural-network-based protection has likewise been explored as a topology-aware learning approach~\cite{kordowich_graph_2025}. Collectively, these studies have shown that \ac{ml} methods can capture temporal patterns and nonlinear relationships in protection-relevant voltage and current signals. At the same time, they have been evaluated under substantially different assumptions regarding topology, fault space, measurement access, preprocessing, timing, and targets.

A central limitation of current practice is its predominantly model-centric perspective. Learning algorithms are usually assessed inside study-specific setups. In each work, the physical system model, data-generation process, measurement configuration, preprocessing pipeline, window construction, target definitions, and validation protocol are specified independently. These choices define the information available to the model, the temporal context available at decision time, and therefore the protection problem actually being solved. As a result, studies that nominally address the same task may in fact evaluate different inference problems. Reported performance is therefore shaped not only by model capability, but also by implicit assumptions about observability, timing, data realism, and system complexity.

Recent survey and scoping studies have confirmed that this is not an isolated issue, but a structural limitation of the field. Reviews of \ac{ml} in power system protection have documented substantial heterogeneity in simulation setups, preprocessing pipelines, feature representations, and evaluation metrics, which limits reproducibility and meaningful cross-study comparison~\cite{vaish_machine_2021,porawagamage_review_2024,kouraichi_deep_2025,oelhaf_scoping_2025}. This echoes the broader reproducibility crisis in machine-learning-based science, where incomplete documentation of data, methods, and evaluation has repeatedly prevented reported results from being reproduced~\cite{gundersen_state_2018}. In a prior scoping review by the authors covering 119 studies, only 16.0\% used real-world data, 82.5\% provided no data access, and only 1.7\% publicly released code or models; even basic metadata such as sampling frequency were omitted in 52.1\% of studies~\cite{oelhaf_scoping_2025}. These omissions are not secondary reporting defects: they determine whether a result can be reproduced, compared, or interpreted under protection-relevant constraints. Practical challenges such as noisy or incomplete measurements, class imbalance, shifts between training and deployment conditions, and limited interpretability further complicate the use of \ac{ml} in safety-critical settings~\cite{porawagamage_review_2024,oelhaf_impact_2025}. These findings indicate that the central limitation is not only a lack of stronger models, but a lack of shared evaluation discipline for turning model results into reproducible protection evidence.

This ambiguity is amplified by the widespread reliance on simulated data~\cite{oelhaf_scoping_2025}. Domain-specific studies targeting \ac{hvdc} systems, wind integration, or hybrid \ac{ac}/\ac{dc} networks provide valuable insights for particular applications~\cite{da_silva_intelligent_2025,uddin_hybrid_2022,yadav_integrating_2025}, but they also introduce additional variation in system modeling, measurement conditions, and operating assumptions. Application-specific studies have further illustrated this dependence on the physical setting, including decision-tree-based fault detection on a \ac{tcsc}-compensated line during power swing~\cite{kumar_mohanty_decision_2023} and an enhanced relaying scheme for a compensated line connected to a \ac{dfig}-based wind farm~\cite{mohanty_enhanced_2024}. When simulation fidelity, scenario coverage, and documentation of the data-generation process are only partly specified, it remains unclear whether reported success or failure is driven by model design, information availability, task formulation, or the realism of the underlying data~\cite{oelhaf_scoping_2025}.

The difficulty is partly interdisciplinary. In protection engineering, the validity of a result depends on physical scope, measurement access, synchronization, timing, operating conditions, and failure behavior. In machine learning, validity depends on data construction, preprocessing, split design, leakage prevention, model selection, and evaluation metrics~\cite{kapoor_leakage_2023,kapoor_reforms_2024}. \ac{ml}-based protection studies require both perspectives simultaneously, but current reporting practices often leave one of them implicit. The resulting gap is procedural: the field needs an evaluation structure that connects protection assumptions with \ac{ml} validity requirements. General-purpose \ac{ai}-assurance standards establish the governance scaffolding but do not supply these protection-specific dimensions; conversely, protection studies rarely adopt \ac{ml} validity safeguards. The framework proposed here is intended to occupy exactly this intersection. Table~\ref{tab:framework_comparison} provides an illustrative structured mapping of how representative protection, machine-learning, and \ac{ai} guidance treats the seven dimensions formalized in the proposed framework.

\begin{table*}[pos=t]
\centering
\caption{Comparison of representative guidance against the seven framework dimensions.}
\label{tab:framework_comparison}
\setlength{\tabcolsep}{3pt}
\begin{tabular*}{\linewidth}{@{\extracolsep{\fill}}lccccccc@{}}
\toprule \textbf{Source family} & \textbf{Obj.} & \textbf{Scope} & \textbf{Obs.} & \textbf{Time} & \textbf{Targets} & \textbf{Validation} & \textbf{Outputs} \\
\midrule
Dataset documentation~\cite{gebru_datasheets_2021} & $\circ$ & $\circ$ & $\circ$ & -- & $\circ$ & $\circ$ & -- \\
ML reporting and validation guidance~\cite{kapoor_reforms_2024,kapoor_leakage_2023,roberts_crossvalidation_2017} & \checkmark & $\circ$ & $\circ$ & $\circ$ & \checkmark & \checkmark & \checkmark \\
Protection reviews and scoping studies~\cite{vaish_machine_2021,porawagamage_review_2024,oelhaf_scoping_2025} & \checkmark & \checkmark & $\circ$ & $\circ$ & $\circ$ & $\circ$ & \checkmark \\
Protection-testing standards~\cite{international_electrotechnical_commission_measuring_2022,international_electrotechnical_commission_measuring_2014} & \checkmark & \checkmark & \checkmark & \checkmark & $\circ$ & $\circ$ & \checkmark \\
General \ac{ai} risk and assurance guidance~\cite{tabassi_artificial_2023} & \checkmark & $\circ$ & $\circ$ & -- & $\circ$ & \checkmark & \checkmark \\
\midrule
\textbf{Proposed framework} & \checkmark & \checkmark & \checkmark & \checkmark & \checkmark & \checkmark & \checkmark \\
\bottomrule \end{tabular*}
\vspace{2pt}
\noindent
\scriptsize
\textit{Note:} This table is an illustrative structured mapping rather than a quantitative score. \checkmark denotes an explicit prescriptive requirement accompanied by an operational procedure, $\circ$ denotes acknowledgement without full operationalization, and -- denotes no identified treatment in the cited sources. Prior-work rows aggregate closely related sources by guidance family.
\end{table*}

The comparison identifies an integration gap rather than an absence of prior guidance: existing source families address complementary subsets of the evaluation problem, whereas the proposed framework operationalizes all seven dimensions jointly. In particular, it treats sample-validity rules, including onset conditioning, as part of the evaluated protection problem rather than as an implicit preprocessing choice; where prior guidance addresses onset at all -- as in the fault-inception-angle definition of \ac{iec}~60255-121 -- it does so as a controlled test parameter, not a sample-validity rule.

Unlike survey and scoping contributions, including prior work by the authors~\cite{oelhaf_scoping_2025}, this paper does not primarily catalog the literature or summarize reported methods. Its contribution is prescriptive rather than descriptive. It defines a standardized evaluation structure, a reporting logic, and a bounded reference instantiation that make study assumptions explicit before model performance is compared.

Overall, the literature has provided substantial evidence of the technical feasibility of \ac{ml} for protection tasks, but it does not yet provide a consistent basis for interpreting why results differ across studies. Addressing this gap requires evaluation frameworks that make task formulation, observability, timing, data provenance, and validation design explicit before model comparison.

\subsection{Objective and Contributions}
\label{sec:intro_contrib}

The objective of this paper is to develop a standardized evaluation framework for \ac{ml}-based power system protection that makes reported results interpretable as conditional and auditable evidence rather than isolated model scores. The framework therefore requires explicit specification of the physical scope, observability, timing constraints, target construction, data provenance, validation design, and reporting outputs before model performance is interpreted or compared. In doing so, it makes the effective inference problem an explicit part of the scientific contribution.

The contributions of this paper are threefold. First, it proposes a standardized, protection-specific evaluation framework that structures the complete path from protection objective and information availability to validation and reporting. The framework defines what must be specified, reported, and examined for an \ac{ml}-based protection result to be reproducible, comparable, and open to independent audit. It thereby provides a first methodological step toward certification-oriented assessment without claiming to constitute a formal certification procedure. Second, it formalizes an interpretation logic in which reported performance is treated as conditional evidence tied to task formulation, physical scope, observability, timing, target construction, and validation design. This separates apparent model capability from differences in the underlying inference problem and prevents aggregate scores from being interpreted independently of the assumptions under which they were obtained. Third, it instantiates the framework in a bounded and reproducible case study on \ac{fc} and \ac{fl}. Under shared sensing, timing, sample-validity, and validation assumptions, the case study demonstrates how the framework exposes task-dependent behavior, observability effects, diagnostic patterns, conventional-reference comparisons, robustness differences, and runtime trade-offs beyond aggregate performance measures. The case study serves as a worked example of the framework; it does not claim benchmark completeness or practical superiority over conventional protection.

The proposed framework complements established protection-testing practice~\cite{international_electrotechnical_commission_measuring_2022} and emerging \ac{ai}-assurance frameworks~\cite{tabassi_artificial_2023,isoiec_information_2023,isoiec_framework_2022} by supplying protection-specific evaluation dimensions, including observability and synchronization, decision horizons, and onset-conditioned sample validity.

\subsection{Paper Organization}
\label{sec:intro_org}

The remainder of this paper is organized as follows. Section~\ref{sec:framework} presents the proposed standardized evaluation framework and its seven-step protocol. Section~\ref{sec:case_study} shows how the framework is instantiated in a bounded case study on a public \ac{emt} benchmark for \ac{fc} and \ac{fl}. Section~\ref{sec:results} reports the resulting findings and illustrates what the framework reveals beyond aggregate performance measures. Section~\ref{sec:discussion} discusses generalizability, reuse, limitations of the present instantiation, and directions for future work. Section~\ref{sec:conclusion} concludes the paper.

\section{Proposed Standardized Evaluation Framework}
\label{sec:framework}

The central contribution of this paper is a standardized evaluation framework for \ac{ml}-based power system protection. The framework does not prescribe a particular model, dataset, or grid architecture. Rather, it specifies the minimum structure a study must define before its results can be interpreted, reproduced, or compared.

The motivation is straightforward: reported performance depends not only on model choice, but also on task definition, physical scope, observability, timing, target construction, data provenance, and validation design. When these elements vary across studies, similar performance numbers may correspond to different underlying inference problems and therefore lack direct comparability. The proposed framework addresses this issue by making the evaluation design explicit, inspectable, and reusable.

Accordingly, the framework structures a study into seven steps: protection objective, physical system scope, observability and measurements, timing and decision windows, targets and valid samples, training and validation protocol, and evaluation outputs and diagnostics. Together, these steps define a standardized reporting structure that makes assumptions explicit and evaluation settings inspectable. The structure is informed by established protection-equipment testing and performance-documentation practice, general \ac{ai} test, evaluation, verification, and validation guidance, and machine-learning reproducibility and temporal-benchmark literature~\cite{international_electrotechnical_commission_measuring_2018,international_electrotechnical_commission_measuring_2014,tabassi_artificial_2023,pineau_improving_2021,kapoor_leakage_2023,wu_current_2022,kim_towards_2022}. Its contribution is not that every dimension is individually unprecedented, but that these previously separate requirements are integrated into one protection-specific and operational evaluation workflow.

\begin{figure*}[pos=t]
\centering
\begin{tikzpicture}[
  x=1cm,
  y=1cm,
  font=\sffamily,
  >={Stealth[length=2.3mm,width=1.6mm]},
  frameworkbar/.style={
    draw=fwink,
    fill=fwink,
    text=white,
    rounded corners=3pt,
    line width=0.8pt,
    minimum width=16.0cm,
    minimum height=10.5mm,
    align=center
  },
  frameworklayer/.style={
    draw=fwedge,
    fill=fwband,
    rounded corners=3pt,
    line width=0.8pt,
    minimum width=16.0cm,
    align=center
  },
  stepcard/.style={
    draw=fwedge!90,
    fill=white,
    rounded corners=2pt,
    line width=0.5pt,
    minimum width=3.65cm,
    minimum height=18.5mm,
    align=center
  },
  stepbadge/.style={
    circle,
    fill=fwink,
    text=white,
    font=\sffamily\bfseries\scriptsize,
    minimum size=5.0mm,
    inner sep=0pt
  },
  headerbadge/.style={
    circle,
    fill=white,
    text=fwink,
    font=\sffamily\bfseries\scriptsize,
    minimum size=5.0mm,
    inner sep=0pt
  },
  layerheading/.style={
    font=\sffamily\fontsize{10.8}{12.4}\selectfont\bfseries,
    text=fwink,
    align=center
  },
  cardtitle/.style={
    font=\sffamily\small\bfseries,
    text=fwink,
    align=center,
    text width=3.05cm
  },
  pipelinebox/.style={
    draw=fwedge!95,
    fill=white,
    rounded corners=1.5pt,
    line width=0.5pt,
    minimum width=2.50cm,
    minimum height=8.0mm,
    align=center,
    font=\sffamily\footnotesize
  },
  phase/.style={
      font=\sffamily\footnotesize\bfseries,
      text=fwink!75,
      align=center
  },
  evidenceitem/.style={
    font=\sffamily\small\bfseries,
    text=fwink,
    align=center,
    text width=3.35cm
  },
  mainflow/.style={
      ->,
      draw=fwink,
      line width=1.35pt,
      >={Stealth[length=3.0mm,width=2.0mm]},
      shorten <=-1.0mm,
      shorten >=-1.0mm
    },
  pipeflow/.style={
    ->,
    draw=fwink,
    line width=0.75pt,
    >={Stealth[length=2.0mm,width=1.4mm]}
  },
  separator/.style={
    draw=fwmid!90,
    densely dotted,
    line width=0.55pt
  }
]

\node[frameworkbar] (stepOne) at (0,0) {};
\node[font=\sffamily\large\bfseries,text=white] at (stepOne.center)
  {DEFINE THE PROTECTION OBJECTIVE};
\node[headerbadge] at ([xshift=5.8mm]stepOne.west) {1};

\node[frameworklayer,minimum height=22.5mm] (specificationLayer) at (0,-1.90) {};

\node[stepcard] (stepTwo)   at (-5.95,-1.90) {};
\node[stepcard] (stepThree) at (-1.98,-1.90) {};
\node[stepcard] (stepFour)  at ( 1.98,-1.90) {};
\node[stepcard] (stepFive)  at ( 5.95,-1.90) {};

\node[stepbadge] at ([xshift=4.3mm,yshift=-4.3mm]stepTwo.north west) {2};
\node[stepbadge] at ([xshift=4.3mm,yshift=-4.3mm]stepThree.north west) {3};
\node[stepbadge] at ([xshift=4.3mm,yshift=-4.3mm]stepFour.north west) {4};
\node[stepbadge] at ([xshift=4.3mm,yshift=-4.3mm]stepFive.north west) {5};

\node at ([yshift=3.6mm]stepTwo.center)
  {\includegraphics[width=8.2mm,height=8.2mm,keepaspectratio]{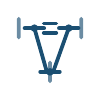}};
\node at ([yshift=3.6mm]stepThree.center)
  {\includegraphics[width=8.2mm,height=8.2mm,keepaspectratio]{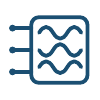}};
\node at ([yshift=3.6mm]stepFour.center)
  {\includegraphics[width=8.2mm,height=8.2mm,keepaspectratio]{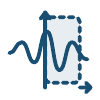}};
\node at ([yshift=3.6mm]stepFive.center)
  {\includegraphics[width=8.2mm,height=8.2mm,keepaspectratio]{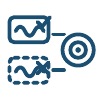}};

\node[cardtitle] at ([yshift=-4.8mm]stepTwo.center)
  {Physical system\\scope};
\node[cardtitle] at ([yshift=-4.8mm]stepThree.center)
  {Observability \&\\measurements};
\node[cardtitle] at ([yshift=-4.8mm]stepFour.center)
  {Timing \& decision\\windows};
\node[cardtitle] at ([yshift=-4.8mm]stepFive.center)
  {Targets \&\\valid samples};

\node[frameworklayer,minimum height=36mm] (protocolLayer) at (0,-5.125) {};
\node[stepbadge] at ([xshift=4.5mm,yshift=-4.5mm]protocolLayer.north west) {6};
\node[layerheading] at ([yshift=-2.4mm]protocolLayer.north)
  {TRAINING AND VALIDATION PROTOCOL};

\node[phase] at (0,-4.02) {Test Phase};

\node[pipelinebox] (testData)  at (-5.60,-4.68) {Data Acquisition};
\node[pipelinebox] (testPrep)  at (-2.80,-4.68) {Preprocessing};
\node[pipelinebox] (testFeat)  at ( 0.0,-4.68) {Feature Extraction};
\node[pipelinebox] (testClass) at ( 2.80,-4.68) {Classification};
\node[pipelinebox] (testPred)  at ( 5.60,-4.68) {Class Prediction};

\path[pipeflow]
  (testData) edge (testPrep)
  (testPrep) edge (testFeat)
  (testFeat) edge (testClass)
  (testClass) edge (testPred);

\draw[separator] (-7.05,-5.32) -- (7.05,-5.32);

\node[pipelinebox] (trainData)  at (-5.60,-5.86) {Data Acquisition};
\node[pipelinebox] (trainPrep)  at (-2.80,-5.86) {Preprocessing};
\node[pipelinebox] (trainFeat)  at ( 0.0,-5.86) {Feature Extraction};
\node[pipelinebox] (trainLearn) at ( 2.80,-5.86) {Model Learning};

\path[pipeflow]
  (trainData) edge (trainPrep)
  (trainPrep) edge (trainFeat)
  (trainFeat) edge (trainLearn);

\draw[pipeflow] (trainLearn.north) -- (testClass.south);
\node[phase] at (0,-6.52) {Training Phase};

\node[frameworklayer,minimum height=18.5mm] (evidenceLayer) at (0,-8.15) {};
\node[stepbadge] at ([xshift=4.5mm,yshift=-4.5mm]evidenceLayer.north west) {7};
\node[layerheading] at ([yshift=-3.5mm]evidenceLayer.north)
  {EVALUATION EVIDENCE};

\draw[fwedge!85,line width=0.45pt] (-4.00,-7.75) -- (-4.00,-8.98);
\draw[fwedge!85,line width=0.45pt] ( 0.00,-7.75) -- ( 0.00,-8.98);
\draw[fwedge!85,line width=0.45pt] ( 4.00,-7.75) -- ( 4.00,-8.98);

\node at (-6.00,-8.10)
  {\includegraphics[width=7.5mm,height=7.5mm,keepaspectratio]{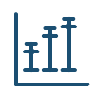}};
\node at (-2.00,-8.10)
  {\includegraphics[width=7.5mm,height=7.5mm,keepaspectratio]{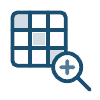}};
\node at ( 2.00,-8.10)
  {\includegraphics[width=7.5mm,height=7.5mm,keepaspectratio]{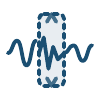}};
\node at ( 6.00,-8.10)
  {\includegraphics[width=7.5mm,height=7.5mm,keepaspectratio]{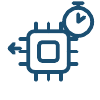}};

\node[evidenceitem] at (-6.00,-8.70) {Task performance};
\node[evidenceitem] at (-2.00,-8.70) {Diagnostics};
\node[evidenceitem] at ( 2.00,-8.70) {Robustness};
\node[evidenceitem] at ( 6.00,-8.70) {Deployment};

\node[frameworkbar,minimum height=10mm] (reportingPackage) at (0,-9.925) {};
\node[font=\sffamily\large\bfseries,text=white] at (reportingPackage.center)
  {STANDARDIZED REPORTING PACKAGE};

\draw[mainflow] (stepOne.south) -- (specificationLayer.north);
\draw[mainflow] (specificationLayer.south) -- (protocolLayer.north);
\draw[mainflow] (protocolLayer.south) -- (evidenceLayer.north);
\draw[mainflow] (evidenceLayer.south) -- (reportingPackage.north);

\end{tikzpicture}
\caption{Seven-step framework for evaluating machine-learning-based power system protection studies. Step~1 defines the protection objective, Steps~2--5 specify the effective inference setting, Step~6 places the classical training and test pipeline within an explicit validation protocol, and Step~7 structures the resulting evaluation evidence. The final reporting package links conclusions to the complete study definition. The pattern-recognition pipeline is adapted from Niemann~\cite{niemann_klassifikation_1983}.}
\label{fig:framework_flowchart}
\end{figure*}
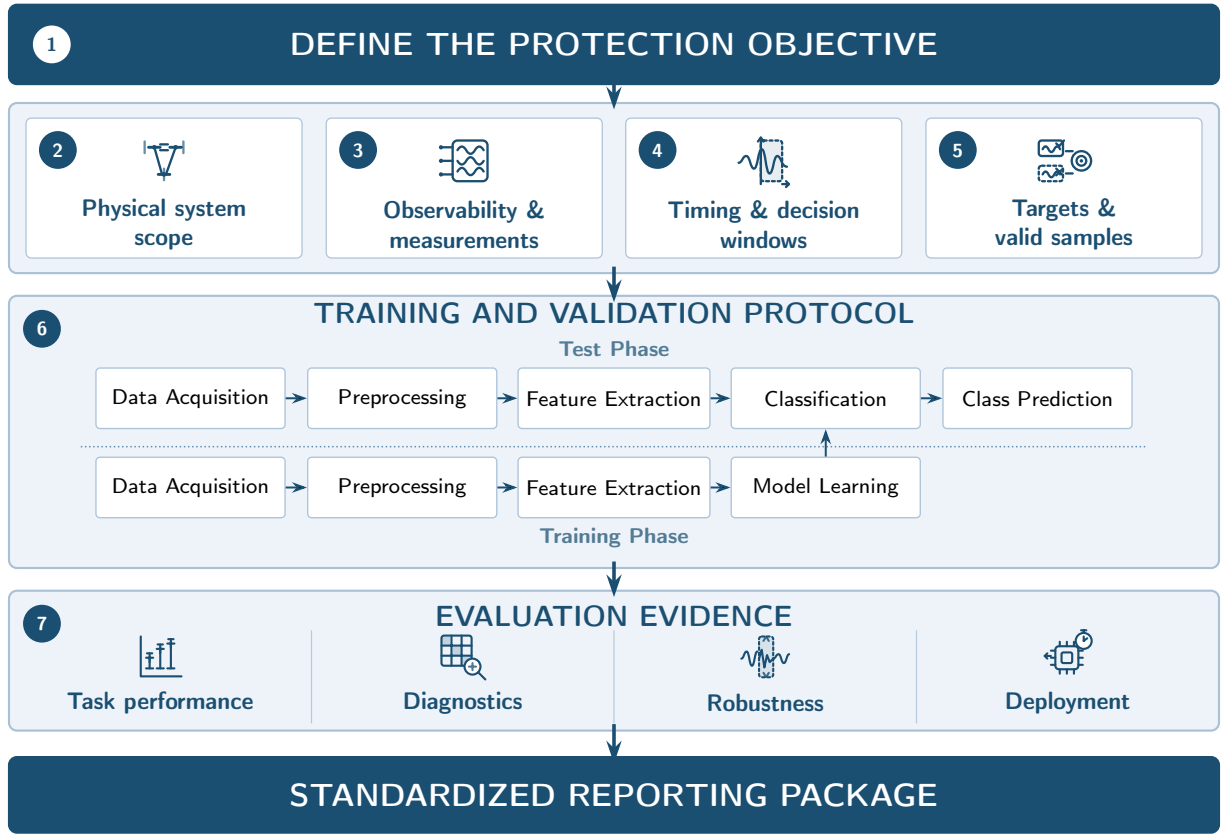

\subsection{Design Goals and Guiding Principles}
\label{sec:framework_principles}

The proposed framework is guided by a small set of principles that define what a rigorous \ac{ml}-based protection study must achieve. First, it must support \textit{comparability}: studies should specify the task, sensing assumptions, timing constraints, targets, and outputs in a form that permits meaningful comparison. Second, it must support \textit{reproducibility}: the full path from raw signals to reported results should be reconstructable, including preprocessing, sample construction, splitting, and reporting. Third, it must remain \textit{physically grounded}: reported performance should be interpretable in the context of topology, operating conditions, fault space, and information availability. Fourth, it must remain \textit{protection-relevant}: evaluation should reflect operational constraints such as short decision times, realistic sensing assumptions, and deployment-oriented decision settings.

In addition, the framework requires \textit{leakage-aware validation}, since related samples can otherwise inflate reported performance. It requires \textit{diagnostic interpretability}, so that evaluation extends beyond aggregate metrics and reveals failure modes across classes, locations, sensing regimes, or operating conditions. Finally, it requires \textit{deployment awareness}, so that predictive quality is assessed together with runtime, sensing requirements, communication assumptions, and robustness under degraded conditions.

Together, these principles define the framework as a reusable evaluation standard for \ac{ml}-based power system protection rather than a benchmark recipe.

\subsection{Framework Overview}
\label{sec:framework_overview}

Figure~\ref{fig:framework_flowchart} summarizes the framework as a seven-step workflow. The sequence runs from problem definition to result interpretation. It first fixes what is to be learned, under which physical and observational conditions, and within which timing constraints. It then specifies how targets are constructed, how training and validation are performed, and which outputs are required to support interpretable conclusions.
The seven steps do not prescribe a particular dataset, topology, model family, or signal domain. Instead, they define the dimensions that must be made explicit in any rigorous protection study. This keeps the framework applicable across different tasks and grid settings while preserving comparability at the level of study design.

The framework output is not only a trained model or a set of performance numbers. It is a standardized reporting package that documents the evaluation setting together with predictive, diagnostic, and deployment-relevant outputs. In this way, the framework supports both reproducible experimentation and interpretation of reported \ac{ml} results relative to explicit assumptions.
The following subsections define each framework step in turn.

\frameworkdivider{Problem Specification}{Steps~1--5 define the task, system, inputs, timing, and prediction target.}

\subsection{Step 1: Protection Objective}
\label{sec:framework_step1}

The first step is to define the protection objective precisely. A study must state which protection function is being addressed, for example, \acl{fd}, \acl{fc}, \acl{fli}, or \acl{fl}, and must specify the corresponding learning formulation, such as binary classification, multi-class classification, or regression.

\fwwhy The protection objective determines what the model is expected to predict, which errors are operationally relevant, and which evaluation outputs are meaningful. The objective must therefore be stated in operational rather than generic \ac{ml} terms. This is consistent with guidance requiring an \ac{ai} system's intended purpose, context of use, expected impacts, and evaluation criteria to be documented before assessment~\cite{tabassi_artificial_2023}. Protection standards similarly specify functions through operational characteristics such as starting, directionality, and time-delay behavior~\cite{international_electrotechnical_commission_measuring_2014}. For example, ``fault analysis'' is too broad to support comparison unless it is decomposed into explicit tasks such as fault detection, fault type classification, or distance estimation. Fixing the protection objective at the outset prevents ambiguity and establishes the basis for later choices such as target construction, metric selection, and diagnostic analysis.

\subsection{Step 2: Physical System Scope}
\label{sec:framework_step2}

Step 2 defines the physical system in which the protection task is posed. A study must specify the network type, topology class, voltage level, operating range, fault and disturbance space, scenario-generation method, coverage of critical operating conditions, data source, and known representativeness limitations. The data source may consist of simulation, hardware-in-the-loop testing, digital-twin-based generation, or field recordings. Together, these choices define both the physical environment from which the data arise and the limits of the protection problem represented by the study.

\fwwhy Task difficulty depends strongly on physical scope. A result obtained on a radial distribution feeder is not directly comparable to a result obtained on a meshed transmission system, even if both are reported for the same nominal task. The same applies to differences in fault and disturbance coverage, scenario-generation assumptions, critical operating conditions, and data origin. Making the physical system scope explicit ensures that reported performance is interpreted relative to the grid conditions under which it was obtained, rather than attributed to the model alone. For simulation-derived studies, established verification and validation practice further distinguishes conceptual-model validity, model verification, operational validity, and data validity~\cite{sargent_verification_2013}.

\subsection{Step 3: Observability and Measurements}
\label{sec:framework_step3}

The third step specifies what information from the physical system is available to the model. A study must specify which measurements are provided, where they are taken, in which signal domain they are represented, how they are synchronized, and whether access is centralized, local, or distributed. This includes, for example, whether the model operates on \ac{emt} waveforms, sampled values, \ac{rms} quantities, phasors, or \ac{pmu}-based features, as well as which voltage, current, frequency, or derived channels are used. These representations are not interchangeable: sampled-value standards define the communication of waveform samples, whereas synchrophasor standards impose explicit time-tagging, synchronization, and static- and dynamic-performance requirements~\cite{international_electrotechnical_commission_communication_2020,international_electrotechnical_commission_measuring_2018}.

A study must also state whether the model receives auxiliary non-waveform information beyond the primary measurements, such as topology encodings, line parameters, bus or line identifiers, operating-point variables, load information, switching states, or other metadata. If channels or auxiliary inputs are missing, delayed, noisy, or otherwise degraded, these conditions must also be reported.

\fwwhy Observability is a primary determinant of task difficulty. Two studies may evaluate the same physical system but still solve different inference problems if one model receives only local measurements while another also receives synchronized multi-location signals, topology information, or operating metadata. Making the available information explicit separates model capability from information availability and turns observability into a comparable evaluation axis rather than a hidden implementation detail. In fault prediction, feature-selection studies provide a complementary way to identify which inputs carry task-relevant information~\cite{kordowich_feature_2026}.

\subsection{Step 4: Timing and Decision Windows}
\label{sec:framework_step4}

The fourth step defines the decision-time setting under which the protection task is evaluated. A study must specify the temporal reference used for sample construction, the decision horizon, the window length, the step size or stride between windows, and any additional real-time constraints that limit what information is available at inference time. This includes the event reference used for alignment, the amount of pre-event and post-event context, the window extraction rule, and whether overlapping windows are used. Together, these choices determine how much pre-fault, fault-onset, and post-fault information the model can use and therefore shape the protection problem being solved.

\fwwhy Protection tasks are time-critical by definition. Protection standards accordingly treat operating-time and time-delay characteristics as explicit quantities to be evaluated under defined test conditions~\cite{international_electrotechnical_commission_measuring_2014}. A result obtained from short windows near fault inception is not directly comparable to one based on longer windows or broader post-event context. Changing the stride also changes the number, overlap, and temporal diversity of the available samples, which can affect both dataset composition and evaluation outcomes. Defining timing and decision windows explicitly ties reported performance to a clear protection-time budget and prevents it from being interpreted independently of the available decision time.

\subsection{Step 5: Targets and Sample Validity}
\label{sec:framework_step5}

The fifth step fixes what the model must predict and which samples are valid for that prediction. A study must specify how labels or regression targets are constructed, which inclusion and exclusion rules are applied, and how boundary cases are handled. This includes, for example, whether samples are labeled by event presence, fault type, faulted line, or fault position, and whether the dataset retains all extracted windows, only windows whose time span overlaps with a fault, or only windows in which the fault begins inside the window. If samples are filtered, merged, discarded, or reassigned, these rules must be stated explicitly.

\fwwhy The task name alone does not fully define the prediction problem. Two studies may both report \ac{fc} or \ac{fl} results while using different class definitions, regression targets, or sample-validity rules. The same applies to boundary handling, such as windows near fault inception, multiple events, or ambiguous cases. Defining targets and sample validity explicitly ensures that reported performance is tied to a fully specified labeling and filtering scheme rather than to labels that are only partly described. More generally, time-series benchmark research has shown that flawed anomaly placement, ambiguous event boundaries, and permissive temporal scoring can materially distort apparent progress and method rankings~\cite{wu_current_2022,kim_towards_2022}.

\frameworkdivider{Evaluation Protocol}{Steps~6--7 define how models are compared and what evidence a study must report.}

\subsection{Step 6: Training and Validation Protocol}
\label{sec:framework_step6}

The sixth step specifies how models are trained, validated, and compared. A study must specify the split design, the unit of independence used for splitting, the treatment of related samples, the preprocessing pipeline, and the conditions under which different models are evaluated. This includes, for example, whether splitting is performed by event, episode, line, feeder, or recording, whether folds are grouped to prevent leakage, and whether preprocessing is fitted on training data only~\cite{roberts_crossvalidation_2017,kapoor_leakage_2023}.

\fwwhy Protection datasets often contain strong dependencies between samples. Multiple windows may originate from the same event, share the same operating point, or overlap in time. If such samples are split across training and test sets, reported performance may reflect leakage rather than generalization. The protocol must therefore state which samples are considered dependent and how this dependency is handled during validation.

It must also define the boundary between training and evaluation. Any normalization, feature extraction, target transformation, dimensionality reduction, or parameter selection must be fitted on training data only and then applied to held-out data without re-estimation. If hyperparameters are tuned, the tuning procedure must be reported explicitly. Competing models must be compared under the same data partitions, target definitions, preprocessing assumptions, and metric definitions so that reported differences are attributable to model behavior rather than to inconsistent evaluation conditions. Reproducibility guidance further emphasizes transparent reporting of data, code, experimental conditions, and model-selection procedures, while benchmark studies show that data sampling, initialization, and hyperparameter choices can materially affect comparative conclusions~\cite{pineau_improving_2021,bouthillier_accounting_2021}.

\subsection{Step 7: Evaluation Outputs and Diagnostics}
\label{sec:framework_step7}

The seventh step defines what outputs a study must report for its results to support clear conclusions. A study must report more than a single headline number. At minimum, the evaluation output should include task-appropriate aggregate metrics, structured diagnostics, and implementation-relevant indicators. Reported uncertainty must identify its source explicitly, for example, variation across held-out groups, model initializations, data samples, or perturbation realizations, because these quantities are not interchangeable~\cite{tabassi_artificial_2023,bouthillier_accounting_2021}. Where repeated evaluations are used, their unit and aggregation procedure must therefore be reported. Depending on the task, this may include classification scores, regression errors, class-resolved confusion patterns, location-resolved error profiles, observability-conditioned comparisons, robustness analyses, and runtime measurements.

\fwwhy Aggregate performance alone does not show why a model succeeds or fails. Two models may achieve similar overall scores while exhibiting different failure modes across fault classes, locations, sensing regimes, or operating conditions. A meaningful evaluation must therefore expose not only average performance, but also the structure and stability of that performance.

The reported outputs must also match the protection objective. Classification tasks require metrics that reflect class-wise behavior and not only overall accuracy. Localization or other regression tasks require error measures with direct physical interpretation and, where relevant, stratified analyses across faulted elements or operating regions. If robustness is assessed, the study must state which factors are varied. If runtime is reported, the measurement setup must be stated clearly enough to support comparison. Recent streaming evaluations further show that short decision windows or early model decisions do not necessarily imply correspondingly short end-to-end protection latency, because the complete signal-processing and inference pipeline contributes additional delay~\cite{abukhousa_latency-aware_2026}. Defining evaluation outputs in this way ensures that a study produces evidence rather than only scores.

\subsection{Standardized Reporting Package}
\label{sec:framework_reporting}

A study that follows the proposed framework should report its evaluation setting in a compact and standardized form, following the broader principle that datasets and evaluation artifacts should be documented for reuse and inspection~\cite{kapoor_reforms_2024,gebru_datasheets_2021}. Just as datasheets document datasets and model cards document trained models for reuse and scrutiny~\cite{mitchell_model_2019}, the proposed reporting package documents the \emph{evaluation setting} of a protection study, so that a result can be inspected, reproduced, and reused under its stated conditions. This reporting package makes the core assumptions of the study explicit and allows readers to judge what is comparable, reproducible, and interpretable. Table~\ref{tab:framework_reporting_template} summarizes the minimum information that should be reported.

\begin{table*}[pos=t]
\centering
\small
\caption{Minimum reporting package for \ac{ml}-based power system protection studies under the proposed framework.}
\label{tab:framework_reporting_template}
\begin{tabular}{p{4.8cm}p{10.3cm}}
\toprule
\textbf{Framework step} & \textbf{Minimum reported information} \\
\midrule
1. Protection objective & Protection task, operational interpretation, learning formulation, and predicted output \\
2. Physical system scope & Network type, topology class, voltage level, operating range, fault and disturbance space, scenario-generation method, critical-condition coverage, data source, and known representativeness limitations \\
3. Observability and measurements & Measurement locations, signal domain, channel set, synchronization assumptions, access regime, and auxiliary inputs \\
4. Timing and decision windows & Event reference, pre-/post-event context, decision horizon, window length, stride, and overlap rule \\
5. Targets and sample validity & Label or regression-target construction, inclusion/exclusion rules, boundary handling, and retained sample set \\
6. Training and validation protocol & Split design, unit of independence, leakage-prevention strategy, preprocessing boundary, and model-comparison conditions \\
7. Evaluation outputs and diagnostics & Aggregate metrics, structured diagnostics, robustness analyses, observability-conditioned results, and runtime setup \\
\midrule
Reporting artifacts & Data/code availability, configuration details, and any restrictions affecting reproducibility or comparison \\
\bottomrule
\end{tabular}
\end{table*}

\section{Case Study: Instantiation on a Public EMT Benchmark}
\label{sec:case_study}

This section instantiates the evaluation framework introduced in Section~\ref{sec:framework} on the public PROTECT-90 \ac{emt} benchmark~\cite{oelhaf_protect-90_2026,kordowich_protect-90_2026}. The case study addresses \ac{fc} and \ac{fl} under shared assumptions on physical system scope, observability, timing, target construction, and validation, as summarized in Table~\ref{tab:case_study_instantiation}. The evaluated methods span conventional protection, classical \ac{ml}, task-specific deep learning, and a pre-trained time-series foundation model, as specified in Sections~\ref{sec:case_study_validation_models} and~\ref{sec:case_study_baselines}. Controlled analyses then vary timing, observability, measurement fidelity, and fault-resistance distribution while keeping the remaining evaluation assumptions fixed. Additional stride and hyperparameter checks assess the stability of the resulting interpretation. Together, these experiments demonstrate how the framework separates model effects from changes in information availability, decision time, data realism, and evaluation configuration.

\begin{table*}[pos=t]
\centering
\small
\caption{Case-study instantiation of the proposed standardized evaluation framework.}
\label{tab:case_study_instantiation}

\begin{tabular}{p{4.6cm}p{10.3cm}}
\toprule
\textbf{Framework step} & \textbf{Reference instantiation} \\
\midrule
1. Protection objective & \ac{fc} and \ac{fl} under shared sensing, timing, and validation assumptions \\
2. Physical system scope & PROTECT-90; 90\,kV double-line topology; 9022 \ac{emt} episodes; randomized fault and operating conditions \\
3. Observability & Synchronized three-phase voltage/current measurements from 8 relays; centralized full-observability reference setting \\
4. Timing & $\pm$80\,ms around fault inception; 20\,ms sliding windows; fixed 5\,ms stride \\
5. Targets & Onset-conditioned 11-class \ac{fc} with one non-onset class and ten fault-type classes; \ac{fl} as normalized fault position; fault-onset-containing windows for \ac{fl} \\
6. Validation & 5-fold grouped cross-validation by simulation episode; shared preprocessing and split design across tasks and models \\
7. Outputs & Aggregate metrics, structured diagnostics, conventional baselines, runtime assessment, and controlled analyses of timing, observability, measurement fidelity, fault-resistance distribution shift, stride, and hyperparameters \\
\bottomrule
\end{tabular}
\end{table*}

\subsection{Physical System and Observability}
\label{sec:case_study_system_observability}

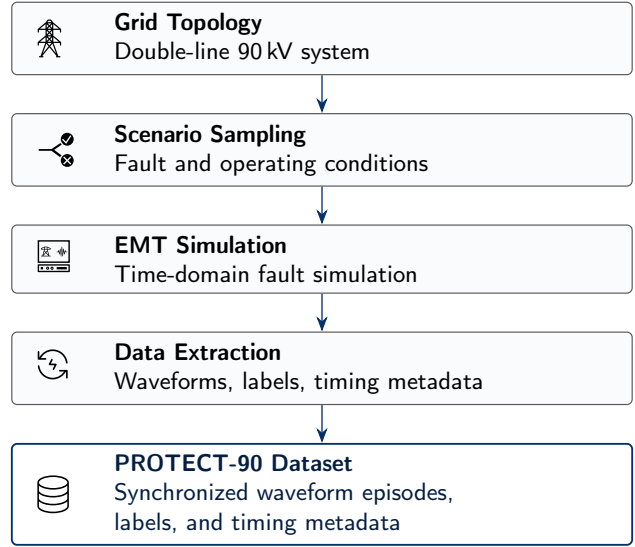
\begin{figure}[pos=t]
\centering

\definecolor{FAUBlue}{RGB}{4,49,106}
\definecolor{FAUDarkBlue}{RGB}{4,30,66}
\definecolor{BoxFill}{RGB}{250,251,253}
\definecolor{LineGray}{RGB}{90,90,90}

\begin{tikzpicture}[
  font=\small,
  node distance=5mm,
  box/.style={
    draw=LineGray,
    fill=BoxFill,
    rounded corners=2pt,
    line width=0.45pt,
    align=left,
    inner xsep=10pt,
    inner ysep=4pt,
    text width=7.5cm,
    minimum height=9mm
  },
  datasetbox/.style={
    draw=FAUBlue,
    fill=white,
    rounded corners=2pt,
    line width=0.7pt,
    align=left,
    inner xsep=10pt,
    inner ysep=4pt,
    text width=7.5cm,
    minimum height=9mm
  },
  arrow/.style={
    -{Stealth[length=2mm]},
    draw=FAUBlue,
    line width=0.5pt
  }
]

\node[box] (topology) {%
  \hspace*{10mm}\textbf{Grid Topology}\\
  \hspace*{10mm}Double-line 90\,kV system};

\node[box, below=of topology] (sampling) {%
  \hspace*{10mm}\textbf{Scenario Sampling}\\
  \hspace*{10mm}Fault and operating conditions};

\node[box, below=of sampling] (simulation) {%
  \hspace*{10mm}\textbf{EMT Simulation}\\
  \hspace*{10mm}Time-domain fault simulation};

\node[box, below=of simulation] (extraction) {%
  \hspace*{10mm}\textbf{Data Extraction}\\
  \hspace*{10mm}Waveforms, labels, timing metadata};

\node[datasetbox, below=of extraction] (dataset) {%
  \hspace*{10mm}\textcolor{FAUDarkBlue}{\textbf{PROTECT-90 Dataset}}\\
  \hspace*{10mm}\textcolor{FAUDarkBlue}{Synchronized waveform episodes,}\\
  \hspace*{10mm}\textcolor{FAUDarkBlue}{labels, and timing metadata}
};

\draw[arrow] (topology.south) -- (sampling.north);
\draw[arrow] (sampling.south) -- (simulation.north);
\draw[arrow] (simulation.south) -- (extraction.north);
\draw[arrow] (extraction.south) -- (dataset.north);

\node at ([xshift=6mm]topology.west)   {\includegraphics[width=5mm]{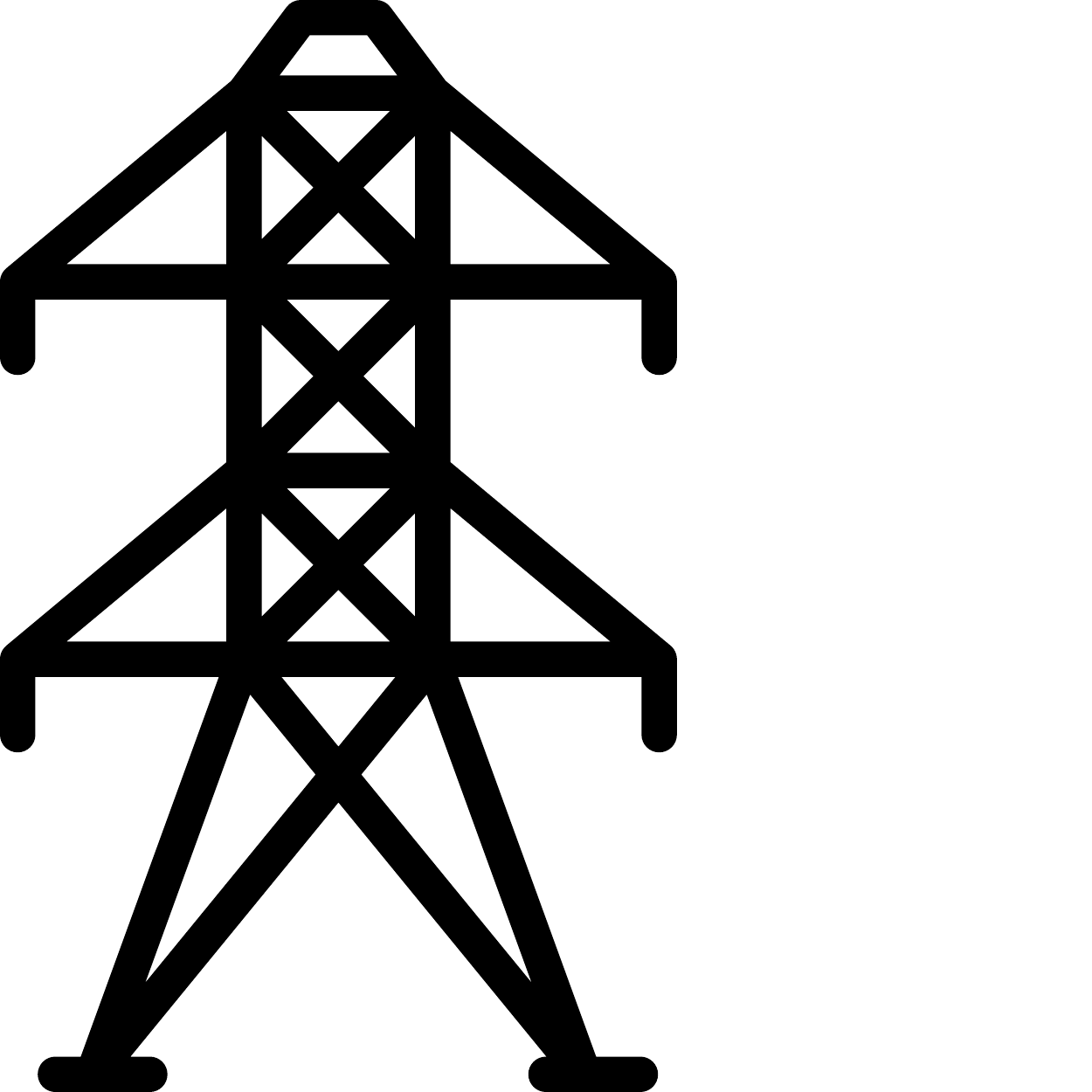}};
\node at ([xshift=6mm]sampling.west)   {\includegraphics[width=5mm]{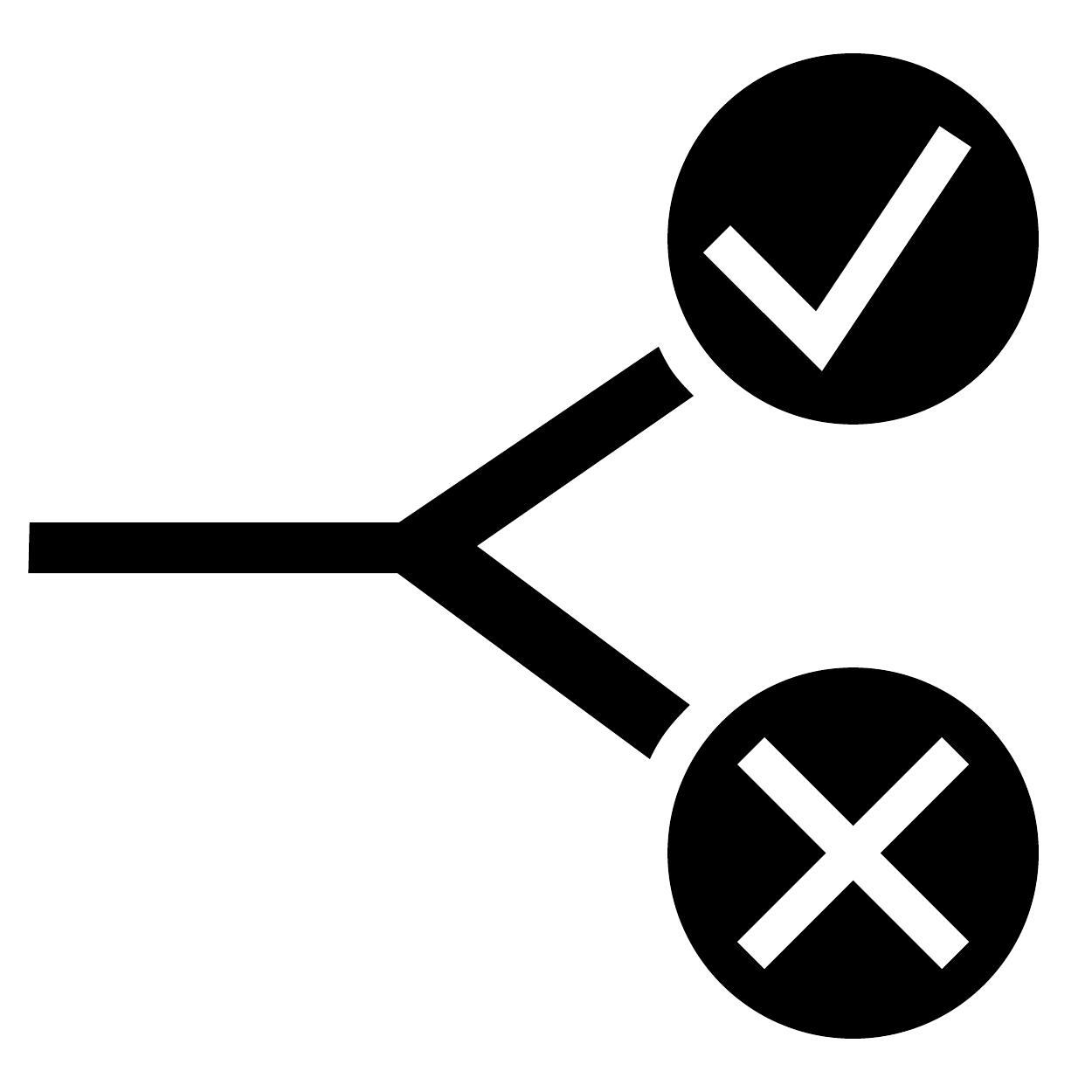}};
\node at ([xshift=6mm]simulation.west) {\includegraphics[width=5mm]{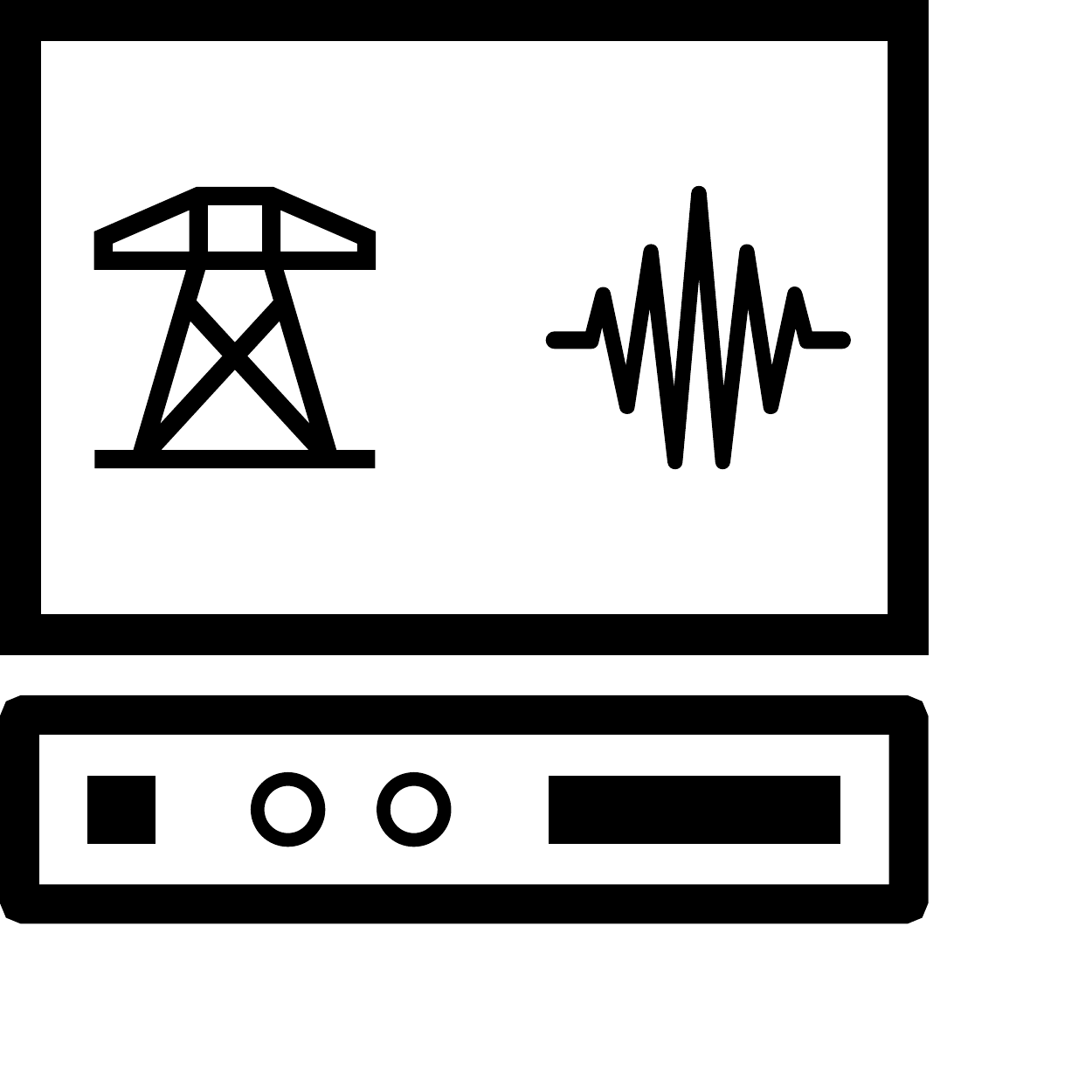}};
\node at ([xshift=6mm]extraction.west) {\includegraphics[width=5mm]{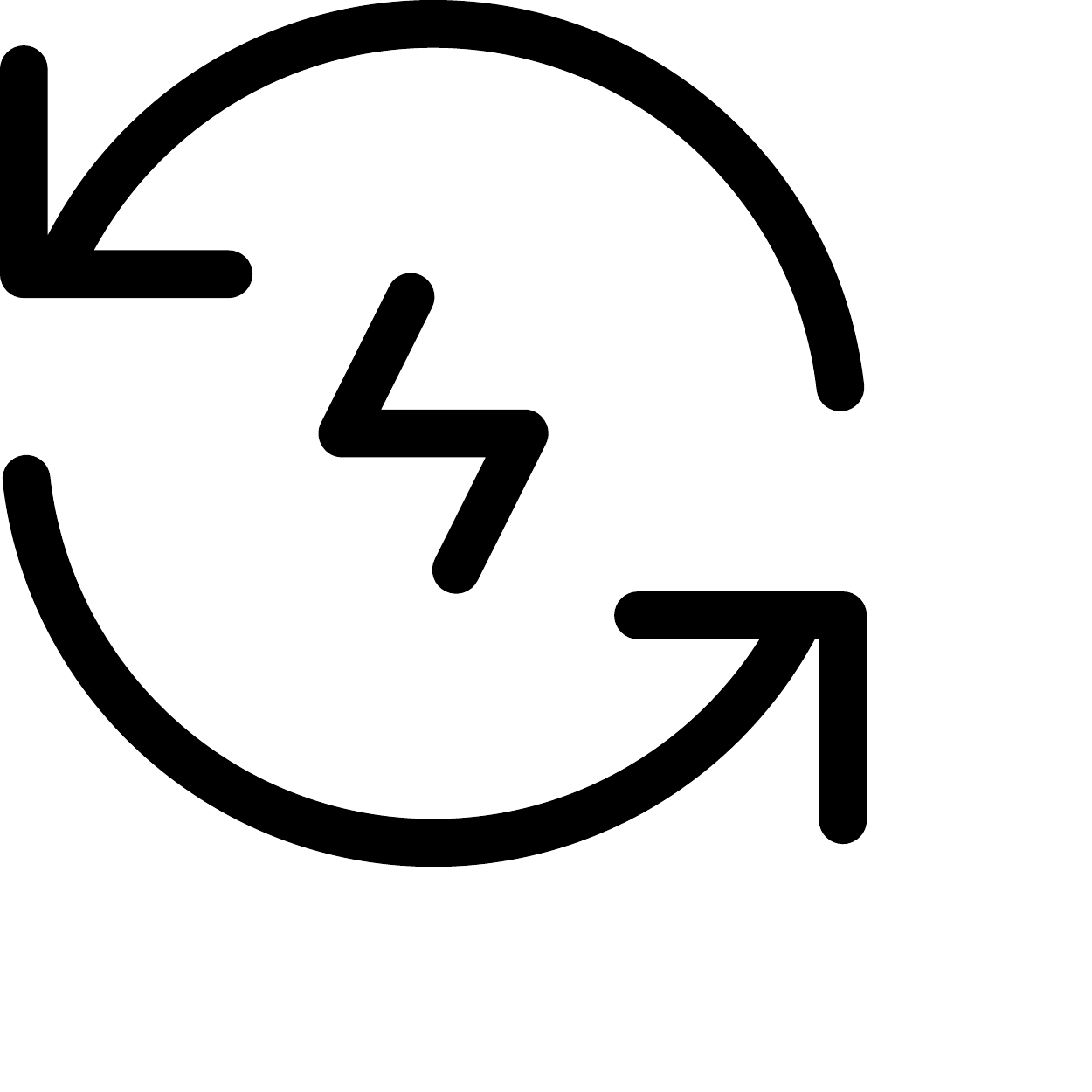}};
\node at ([xshift=6mm]dataset.west)    {\includegraphics[width=5mm]{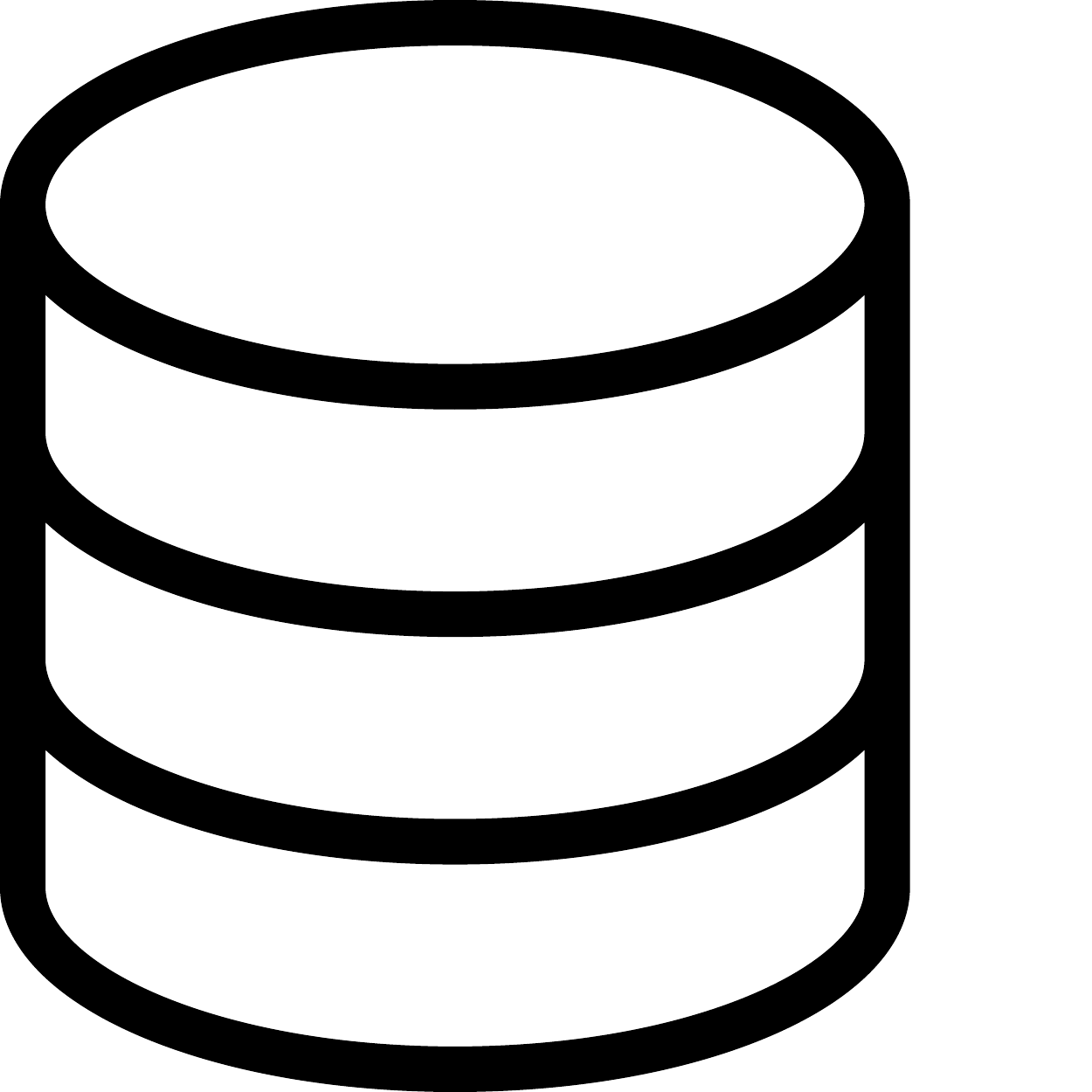}};

\end{tikzpicture}

\caption{Case-study data generation flow for the PROTECT-90 dataset~\cite{oelhaf_protect-90_2026}. A fixed grid topology is combined with randomized fault and operating scenarios, simulated in the \ac{emt} domain, and processed into synchronized waveform episodes with labels and metadata.}
\label{fig:data_generation_flow}
\end{figure}

The case study is based on the public PROTECT-90 dataset~\cite{oelhaf_protect-90_2026}. The benchmark is derived from a 90\,kV transmission-system ``Double Line'' topology with multiple relay locations and diverse short-circuit events. All episodes are generated in DIgSILENT PowerFactory using the \ac{emt} simulation module. This simulation-based design reflects the current scarcity of publicly accessible real-world protection datasets with sufficiently detailed waveform recordings and labels for reproducible \ac{ml} evaluation~\cite{oelhaf_scoping_2025,wilson_grid_2024,evdakov_dataset_2026,gillioz_large_2025}. These public resources provide valuable event signatures, real-world oscillograms, or synthetic transmission-grid data, but differ in signal representation, measurement layout, available metadata, and supported protection targets and therefore do not constitute directly interchangeable benchmarks for the present \ac{fc}/\ac{fl} protocol. Scenario diversity is introduced through domain randomization over fault and operating parameters, including variations in fault resistance, inception time, line characteristics, loading conditions, and external grid properties within predefined physically plausible ranges~\cite{oelhaf_protect-90_2026,wang_generic_2022}. Figure~\ref{fig:data_generation_flow} summarizes the corresponding data-generation process. Such simulation-based benchmarks require explicit documentation of the data-generation assumptions because synthetic power-system datasets can otherwise be difficult to interpret or validate across studies~\cite{krishnan_validation_2020}.

The resulting benchmark contains 9022 simulation episodes. Each episode spans $T = 1\,\mathrm{s}$ and is sampled at $f_s = 6400\,\mathrm{Hz}$, yielding 6400 time steps per episode. The simulated events cover the main short-circuit categories considered in this study, namely \ac{slg}, \ac{ll}, \ac{llg}, and \ac{lll}.

The measurement model is based on primary three-phase voltage and current waveforms extracted at protection-relevant locations. In total, each episode provides signals from $N_{\mathrm{PR}} = 8$ locations corresponding to the relay positions in the studied topology. At each location, three-phase voltage and three-phase current signals are available, resulting in six channels per location. Under the centralized reference configuration, all available measurements are jointly provided to the learning model, yielding the episode-level representation

\begin{equation}
\centering
X \in \mathbb{R}^{N_{\mathrm{PR}} \times 6 \times T_s},
\label{eq:case_study_input_tensor}
\end{equation}

where $T_s = 6400$ denotes the number of discrete time steps per episode. No auxiliary non-waveform metadata are used in the reference configuration. All measurements are treated as fully synchronized, noise-free, and continuously available, with no communication delay, jitter, or packet loss.

This centralized full-observability configuration serves as an upper-bound reference sensing setting for the case study. Later analyses vary the available measurements by restricting the input either to a single relay location or to both terminals of the same line, while preserving the remaining study design. These reduced-observability experiments therefore act as controlled sensitivity analyses of the observability dimension rather than as separate case-study definitions. The clean reference configuration excludes non-fault disturbances, \ac{ct}/\ac{vt} nonidealities, missing channels, synchronization errors, and communication effects, all of which are relevant for protection-equipment assessment and practical relay evaluation~\cite{international_electrotechnical_commission_measuring_2022}. Additive noise, a simplified current-transformer saturation proxy, and synchronization jitter are subsequently evaluated as controlled measurement-fidelity axes in Section~\ref{sec:results_fidelity}. Sensor-availability degradation is addressed in the companion study~\cite{oelhaf_robustness_2026}, whereas communication latency and broader non-fault disturbances remain outside the present scope.

\subsection{Timing, Windowing, and Targets}
\label{sec:case_study_timing_targets}

Sample construction is defined relative to the fault inception time $t_{\mathrm{f}}$. For each episode, the raw tensor in Eq.~\ref{eq:case_study_input_tensor} is restricted to a $\pm$ 80\,ms interval around $t_{\mathrm{f}}$, retaining a short pre-fault baseline together with the early post-fault transient. This bounded crop is used as a reference timing setting that preserves protection-relevant onset information without introducing substantially longer post-event context that would change the effective decision problem. Importantly, this cropping uses the ground-truth $t_{\mathrm{f}}$ from simulation metadata, which is unavailable pre-detection in a deployed relay. The reference configuration therefore isolates \ac{fc} and \ac{fl} from the upstream detection problem; a deployment-oriented instantiation would require either a coupled detection stage or a trigger derived from observable signal features alone.

From this cropped segment, overlapping sliding windows are extracted. Let $L$ denote the window length in samples and $S$ the step size between consecutive windows. The $i$-th window is
\begin{equation}
x_i = X[:, :, iS : iS+L],
\end{equation}
with start time $t_i = iS/f_s$.

The reference timing configuration uses a fixed step size of $S = 32$ samples ($5\,\mathrm{ms}$) and a fixed window length of $L = 128$ samples, corresponding to a decision horizon of $20\,\mathrm{ms}$. At a nominal system frequency of 50\,Hz, this window covers approximately one fundamental cycle and therefore provides a physically interpretable reference horizon for protection-oriented evaluation. The 5\,ms stride provides overlapping temporal views of the same transient while retaining bounded sample density; it therefore affects both sample overlap and the effective composition of the resulting dataset. For this reason, stride is treated as part of the evaluation setting rather than as a neutral implementation detail. Additional window lengths are examined later as a controlled sensitivity analysis of the timing dimension.

Both tasks are defined on this shared windowed representation. For \ac{fc}, each window receives a discrete label
\begin{equation}
y_{\mathrm{FC}} \in \{c_0, c_1, \dots, c_{10}\},
\end{equation}
where $c_1$--$c_{10}$ correspond to the short-circuit classes AG, BG, CG, AB, BC, CA, ABG, BCG, CAG, and ABC. A window receives one of these fault-type labels only when the fault inception lies within the window; all other windows, including pre-fault windows and windows in which the fault is already active, receive $c_0$. Thus, $c_0$ denotes the absence of a fault inception within the current window rather than a physically fault-free state.

For \ac{fl}, the target is the normalized fault position along the affected line segment,
\begin{equation}
y_{\mathrm{FL}} = \frac{d_{\mathrm{fault}}}{\ell_{\mathrm{line}}} \cdot 100,
\end{equation}
where $d_{\mathrm{fault}}$ is the distance from a fixed reference terminal and $\ell_{\mathrm{line}}$ is the line length. Reporting the target in percent of line length makes localization results comparable across line segments.

For \ac{fl}, only windows containing the fault onset are retained. A window $x_i$ is used only if
\begin{equation}
t_i + \epsilon < t_{\mathrm{f}} < t_i + \frac{L}{f_s} - \epsilon,
\end{equation}
where $\epsilon = 2/f_s$ is a two-sample margin. This excludes boundary cases and ensures that each retained regression sample contains both pre-fault information and the initial post-fault transient. This onset-conditioned definition is an intentional task-construction choice that focuses localization on early transient evidence rather than on broader post-fault context. It should therefore be interpreted as one bounded formulation of \ac{fl}, not as the only valid definition of the localization problem.

\subsection{Validation Protocol and Model Families}
\label{sec:case_study_validation_models}

The reference, timing, observability, conventional-baseline, measurement-fidelity, stride, and hyperparameter analyses use five-fold cross-validation grouped by simulation episode. The fault-resistance distribution-shift analysis instead uses the disjoint episode-level training and test partitions defined in Appendix~\ref{app:generalization}. All windows originating from the same episode are assigned to the same fold. This grouping is necessary because sliding windows from the same episode are strongly dependent through shared waveform structure, fault metadata, and operating conditions. Grouped splitting therefore prevents leakage that would otherwise inflate reported performance~\cite{roberts_crossvalidation_2017,kapoor_leakage_2023}. Within each investigation, all compared models use the same deterministic episode-grouped split, task definition, timing configuration, observability setting, and metric definitions. The split is also reused across investigations wherever the underlying sample construction is unchanged. Unless stated otherwise, reported values are the mean and standard deviation across the five held-out fold scores. These fold-wise standard deviations describe variation across episode groups and do not quantify variation across model initializations, confidence intervals, or standard errors.

For the classical \ac{ml} estimators, each extracted window is converted into a fixed-length feature vector by reshaping the windowed tensor to
\begin{equation}
x_i \in \mathbb{R}^{L \times F}
\;\mapsto\;
\tilde{x}_i \in \mathbb{R}^{L \cdot F},
\end{equation}
where \(L\) denotes the number of time steps in the window and \(F = N_{\mathrm{PR}} \cdot 6\) the number of input features per time step. In implementation terms, this corresponds to \texttt{reshape(n\_samples, n\_timesteps * n\_features)}, that is, a time-major flattening in which all features at one time step are kept together before proceeding to the next. For each fold, preprocessing consists of standardizing features to zero mean and unit variance, with the transformation fitted on the training partition only and then applied unchanged to the corresponding held-out fold. Hyperparameters are fixed across folds and selected without access to test-fold results; no fold-specific retuning is performed. The \ac{cnn1d} receives the same fold-wise standardized waveform windows as the classical models, but the standardized arrays are reshaped to channels-by-time rather than flattened for prediction. For each outer fold, the standardizer is fitted on the complete outer training partition only and then applied unchanged to the held-out fold. The fixed, untuned architecture comprises three one-dimensional convolution blocks with 64, 128, and 128 output channels; each block uses a kernel width of 5, batch normalization, rectified-linear activation, and max pooling by a factor of 2. Global average pooling and a linear output layer produce the task prediction. Models are trained with Adam at a learning rate of $10^{-3}$ and batch size 256 for at most 50 epochs. Early stopping with patience 8 uses a 10\% episode-grouped validation subset of the outer training fold. Cross-entropy loss is used for \ac{fc} and mean-squared-error loss for \ac{fl}. MOMENT-1-large is used as a frozen pre-trained time-series encoder, with its representations read out by either a linear probe or a two-layer \ac{mlp} prediction head with approximately 25.4 million trainable parameters. The input-length adaptation, channel-wise encoding and embedding aggregation, linear-probe solver and regularization, \ac{mlp}-head dimensions and activation, optimizer, learning rate, batch size, epoch limit, task-specific loss, validation split, and stopping criterion are fixed in the released experiment configuration and reused unchanged across folds. For MOMENT, the fold-local protocol is applied end to end: within every outer fold, the raw-window standardizer is fitted on the outer-training windows only, the training and held-out embeddings are regenerated independently using that fold's standardizer and the frozen encoder, and the linear probe and \ac{mlp} head are fitted without any access to the held-out fold. These models use the same episode-grouped outer folds, task definitions, decision horizons, and task-specific metrics as the classical learning models; the \ac{cnn1d} results are reported in Section~\ref{sec:results_transformer} and the MOMENT results in Appendix~\ref{app:transformer}.

The learned-model evaluation uses three complementary model panels. The broad classical panel comprises \Ac{ridge}, \Ac{knn}, \Ac{gb}, and \Ac{mlp} estimators, together with their corresponding regression variants for \ac{fl}, and is used for the reference and timing analyses. The focused sensitivity panel comprises the neural \ac{mlp} and histogram-based \ac{gb}, which provide contrasting nonlinear estimators for the observability, measurement-fidelity, fault-resistance-shift, runtime, stride, and hyperparameter analyses. The cross-family panel comprises the \ac{mlp}, a task-specific \ac{cnn1d}, and MOMENT-1-large, with the \ac{mlp} providing continuity with the preceding analyses. The classical models are implemented in \texttt{scikit-learn} 1.8.0~\cite{pedregosa_scikit-learn_2011}. Their fixed configurations are summarized in Table~\ref{tab:fixed_model_configurations}, while the separate single-parameter sensitivity analyses are reported in Appendix~\ref{app:robustness}. Conventional protection methods are evaluated separately in Section~\ref{sec:case_study_baselines}.

\subsection{Conventional Protection Baselines}
\label{sec:case_study_baselines}
Conventional protection methods are evaluated under the same episode-grouped five-fold splits and task-specific metrics as the learning models. The per-window localization comparison uses the onset-conditioned windows defined in Section~\ref{sec:case_study_timing_targets}. Because the phasor-based locators require settled post-onset quantities, separate per-episode settled estimates are additionally reported and treated as a distinct sample-validity setting rather than as validity-matched learning-model results. All methods use phasors obtained from a full-cycle discrete Fourier transform at 50\,Hz, following standard digital-relaying and distance-protection practice~\cite{phadke_computer_2009,ziegler_numerical_2011}. No additional signal pre-filtering is applied before phasor estimation, which should be considered when interpreting the conventional results relative to production relay implementations. At $f_s=6400$\,Hz, a 20\,ms window contains exactly 128 samples and therefore one nominal cycle. Conventional baselines are consequently evaluated at the 20\,ms and 50\,ms horizons; the 10\,ms horizon is marked phasor-invalid.

For \ac{fc}, a rule-based symmetrical-component phase selector determines ground involvement from the residual-current ratio $|3I_0|/|I_1|$, where $I_0$ and $I_1$ denote the zero- and positive-sequence currents, and identifies faulted phases from their current increase relative to a pre-fault reference, following established residual- and phase-current selection principles~\cite{blackburn_protective_2014,phadke_computer_2009}. The resulting phase set and ground flag are mapped to the same eleven classes used for the learning models. The phase-pickup threshold $\tau_p$ and ground-involvement threshold $\tau_g$ are selected by grid search on the training folds using macro-\ac{f1} and are frozen for the held-out fold, thereby placing conventional parameter selection within the Step-6 training boundary.

For \ac{fl}, an uncompensated single-ended reactance-based locator, grounded in established one-terminal fault-location formulations~\cite{takagi_development_1982,saha_fault_2010}, and a synchronized two-ended positive-sequence locator, following established two-terminal fault-location principles~\cite{johns_accurate_1990,ieee_power_and_energy_society_ieee_2015}, represent single-relay and relay-pair observability, respectively. Both receive the ground-truth faulted line and fault type for loop selection, matching the isolated \ac{fl} formulation used for the learning models. They additionally use the episode-specific line length and sequence impedances from the benchmark metadata and therefore operate under best-case parameter observability. By contrast, the centralized learning models use waveform inputs only. Estimated distance is expressed as a percentage of line length from the lower-index bus. Results are reported per window and, for the conventional locators, per episode using the settled post-onset estimate.

\subsection{Measurement-Fidelity Non-Idealities}
\label{sec:case_study_fidelity}
To address the realism of the reference configuration, three measurement-fidelity non-idealities are introduced as controlled, one-axis-at-a-time sensitivity analyses, mirroring the existing timing and observability sweeps. Additive white Gaussian noise is applied per channel and per window at a target signal-to-noise ratio; a proxy for current-transformer saturation magnitude-clips the current channels at a fraction of their clean peak (a deliberately simple, symmetric proxy; a faithful flux-driven current-transformer model~\cite{ieee_ieee_2023} is designated future work); and synchronization jitter shifts each relay's channels by an independent integer sample offset. Sensor-\emph{availability} degradation (missing channels, downsampling, communication dropout) is not reproduced here, as it is the subject of a separate published study on the same data~\cite{oelhaf_robustness_2026}; the present analysis adds the complementary measurement-\emph{fidelity} axes that study does not cover.

Perturbations are injected on the raw waveform windows before standardization. The primary protocol is train-clean and test-perturbed: the clean fold models are trained once and evaluated on perturbed held-out folds, with the standardizer kept fit on clean training statistics, mirroring a system calibrated under nominal conditions that then meets degraded inputs. Each stochastic level is repeated over five realizations seeded deterministically from a single global seed. For each fold and perturbation level, the metric is first averaged across the five realizations; reported values are then computed as the mean and standard deviation across the five fold-level means. The clean level of every axis is the identity operator and therefore reproduces the corresponding reference means. The clean rows report the corresponding five-fold reference statistics; perturbed levels are aggregated over folds and realizations as stated in the table captions.

\subsection{Reproducibility}
\label{sec:case_study_repro}

Except for the explicitly multi-seed fault-resistance-shift experiment, all experiments use a global pseudorandom seed of 42, with deterministic derivation of perturbation seeds for each combination of fold, evaluation axis, level, and realization. The episode-grouped five-fold split is deterministic and reused across tasks, models, decision horizons, and observability settings. Learning-model hyperparameters are fixed a priori and are not selected using held-out benchmark results. The ablations reported in Appendix~\ref{app:robustness} are diagnostic sensitivity analyses rather than part of the model-selection procedure. For the conventional phase selector, thresholds are selected independently within each training fold and then frozen for evaluation on the corresponding held-out fold.

The classical-model environment is pinned to \texttt{scikit-learn} 1.8.0, and the released configuration records all explicitly set estimator parameters. The \ac{cnn1d} runs used PyTorch 2.8.0 with CUDA 12.8; their logs record the software version, pseudorandom seed, fold-wise stopping epoch, and GPU model. NumPy and PyTorch pseudorandom seeds are fixed to 42, although bitwise-identical execution across different CUDA devices is not claimed. The released code also records the configuration and hardware used for runtime measurements. The \ac{fc} reference, timing, observability, runtime, and hyperparameter-ablation experiments, including the 10\,ms configurations, are regenerated under this pinned environment. The \ac{knn} and \ac{ridge} baselines remain unchanged, while the regenerated \ac{mlp} and histogram-based \ac{gb} results differ from the earlier reported values by at most $0.008$ in macro-\ac{f1}. All regenerated estimators use the fixed pseudorandom seed recorded in the released configuration. The released code and configurations support regeneration of the reported experiments subject to the documented data and dependency-access requirements.

\subsection{Reported Outputs}
\label{sec:case_study_outputs}

Consistent with the reporting logic defined in Section~\ref{sec:framework}, the case study reports aggregate predictive performance, structured diagnostics, conventional-method context, runtime measurements, and controlled analyses of timing, observability, measurement fidelity, fault-resistance distribution shift, stride, and hyperparameter sensitivity. For \ac{fc}, the primary aggregate metric is macro-\ac{f1}, complemented by confusion-based diagnostics that make class-resolved behavior explicit, including confusions between the non-onset class and the onset-conditioned fault-type classes. The classification target is strongly imbalanced -- the non-onset class accounts for $89.7\%$ of windows and the rarest fault-type class for $0.83\%$ (an ${\approx}108{:}1$ ratio; see Table~\ref{tab:class_distribution}) -- so macro-averaged \ac{f1} is reported rather than accuracy, since a trivial non-onset-only classifier attains $89.7\%$ accuracy but only ${\approx}0.09$ macro-\ac{f1}. For \ac{fl}, the primary error measure is \ac{mae}. Runtime measurements refer to model-side inference on pre-windowed inputs under the stated reference assumptions. Compact robustness analyses are included to assess whether the main conclusions remain stable under reasonable parameter variation.

Together, these outputs define the evidence reported for the instantiated study. Section~\ref{sec:results} uses them to analyze the empirical findings of the reference configuration and to show how controlled variations of selected evaluation dimensions affect the resulting conclusions.

\section{Results and Framework-Guided Interpretation}
\label{sec:results}

This section reports the empirical findings of the bounded reference case study introduced in Section~\ref{sec:case_study}. The purpose is not to present benchmark scores in isolation, but to show what becomes interpretable once task definition, physical scope, observability, timing, targets, and validation are fixed explicitly. The results are therefore read as evidence for the framework rather than as a model-ranking exercise.

The section begins with the fixed 20\,ms reference configuration and then examines timing and observability as controlled problem-definition axes. Conventional methods are subsequently placed within the same reporting structure to expose the role of sensing and auxiliary information. Measurement-fidelity degradation and fault-resistance distribution shift then test robustness beyond the clean in-distribution setting, while structured diagnostics, runtime measurements, stride sensitivity, and hyperparameter analyses examine the stability and practical interpretation of the reported results. Finally, a cross-family comparison of the \ac{mlp} and \ac{cnn1d} (with a pre-trained foundation-model baseline in Appendix~\ref{app:transformer}) tests whether the task-level interpretation persists beyond the classical model panel.

\begin{table*}[pos=t]
\caption{Reference performance across representative case-study models for the fixed 20\,ms timing configuration. \ac{fc} is reported by macro-\ac{f1}; \ac{fl} by \ac{mae} in percent of normalized line length. Values are mean $\pm$ std over 5 folds.}
\small
\label{tab:reference_20ms}
\centering
\begin{tabular}{llcccc}
\toprule
\bfseries Task & \bfseries Metric & \bfseries \acs{mlp} & \bfseries \ac{gb} & \bfseries \ac{knn} & \bfseries Ridge \\
\midrule
\ac{fc} & Macro-\ac{f1} $\uparrow$ & \textbf{0.991 $\pm$ 0.001} & 0.745 $\pm$ 0.017 & 0.799 $\pm$ 0.006 & 0.095 $\pm$ 0.003 \\
\ac{fl} & \ac{mae} [\%] $\downarrow$ & \textbf{10.20 $\pm$ 0.25} & 14.78 $\pm$ 0.25 & 20.10 $\pm$ 0.16 & 26.12 $\pm$ 0.36 \\
\bottomrule
\end{tabular}
\end{table*}

\subsection{Reference Performance Under the Fixed 20\,ms Configuration}
\label{sec:results_reference_20ms}

Table~\ref{tab:reference_20ms} summarizes the reference results for the fixed 20\,ms configuration defined in Section~\ref{sec:case_study}. Because \ac{fc} and \ac{fl} are evaluated under the same physical scope, observability setting, temporal budget, and validation protocol, the comparison isolates task-dependent differences rather than differences induced by the evaluation setup.

For \ac{fc}, the reference results indicate that short-window fault classification is already highly effective under the present benchmark and sensing assumptions. The \ac{mlp} attains a macro-\ac{f1} of $0.991 \pm 0.001$, whereas the remaining representative models perform substantially worse. The weak performance of the linear \ac{ridge} baseline further suggests that the classification task is strongly non-linear in the present representation.

For \ac{fl}, the pattern is different. Although the \ac{mlp} again yields the strongest result, the lowest error remains an \ac{mae} of 10.20 $\pm$ 0.25\,\% of normalized line length, with \ac{gb}, \ac{knn}, and \ac{ridge} producing progressively larger errors. Under the same benchmark, observability setting, and temporal budget, classification approaches its metric ceiling, whereas localization retains physically material error. The fixed 20\,ms reference configuration therefore reveals a task-dependent performance asymmetry under otherwise shared evaluation assumptions.

\begin{table*}[pos=t]
\caption{Controlled timing sensitivity for fault classification. Macro-\ac{f1} across decision horizons for the representative case-study models. Mean $\pm$ std over 5 folds; higher is better. The 20\,ms row corresponds to the fixed reference timing configuration.}
\small
\label{tab:fc_timing_sensitivity}
\centering
\begin{tabular}{lcccc}
\toprule
\bfseries Window & \bfseries \ac{mlp} & \bfseries \ac{gb} & \bfseries \ac{knn} & \bfseries \ac{ridge} \\
\midrule
10 ms & \textbf{0.985 $\pm$ 0.011} & 0.418 $\pm$ 0.009 & 0.738 $\pm$ 0.012 & 0.097 $\pm$ 0.002 \\
20 ms (ref.) & \textbf{0.991 $\pm$ 0.001} & 0.745 $\pm$ 0.017 & 0.799 $\pm$ 0.006 & 0.095 $\pm$ 0.003 \\
30 ms & \textbf{0.988 $\pm$ 0.005} & 0.977 $\pm$ 0.001 & 0.831 $\pm$ 0.004 & 0.093 $\pm$ 0.002 \\
40 ms & \textbf{0.988 $\pm$ 0.007} & 0.982 $\pm$ 0.002 & 0.851 $\pm$ 0.003 & 0.091 $\pm$ 0.002 \\
50 ms & \textbf{0.990 $\pm$ 0.005} & 0.982 $\pm$ 0.001 & 0.863 $\pm$ 0.003 & 0.089 $\pm$ 0.002 \\
\bottomrule
\end{tabular}
\end{table*}

\begin{table*}[pos=t]
\caption{Timing sensitivity for \ac{fl}. \ac{mae} is reported as percent of normalized line length across decision horizons. Values are mean $\pm$ std over 5 folds; lower is better. The 20\,ms setting is the reference configuration.}
\small
\label{tab:fl_timing_sensitivity}
\centering
\begin{tabular}{lcccc}
\toprule
\bfseries Window & \bfseries \ac{mlp} & \bfseries \ac{gb} & \bfseries \ac{knn} & \bfseries \ac{ridge} \\
\midrule

10 ms & \textbf{10.64 $\pm$ 0.36} & 14.65 $\pm$ 0.19 & 20.38 $\pm$ 0.32 & 26.14 $\pm$ 0.32 \\
20 ms (ref.) & \textbf{10.20 $\pm$ 0.25} & 14.78 $\pm$ 0.25 & 20.10 $\pm$ 0.16 & 26.12 $\pm$ 0.36 \\
30 ms & \textbf{10.18 $\pm$ 0.35} & 14.64 $\pm$ 0.18 & 19.65 $\pm$ 0.19 & 26.11 $\pm$ 0.35 \\
40 ms & \textbf{10.46 $\pm$ 0.52} & 14.59 $\pm$ 0.19 & 19.33 $\pm$ 0.16 & 26.12 $\pm$ 0.32 \\
50 ms & \textbf{9.92 $\pm$ 0.27} & 14.66 $\pm$ 0.20 & 19.12 $\pm$ 0.14 & 26.13 $\pm$ 0.31 \\
\bottomrule
\end{tabular}
\end{table*}

\subsection{Controlled Timing Sensitivity}
\label{sec:results_timing_sensitivity}

Tables~\ref{tab:fc_timing_sensitivity} and~\ref{tab:fl_timing_sensitivity} summarize the effect of varying the decision horizon while keeping the remaining case-study design fixed. In framework terms, this analysis isolates Step~4 and examines how the two protection objectives depend on available decision time under otherwise identical assumptions.

For \ac{fc}, Table~\ref{tab:fc_timing_sensitivity} shows that performance is most sensitive at the shortest horizons and then approaches saturation. The \ac{mlp} already attains a macro-\ac{f1} of $0.985 \pm 0.011$ at 10\,ms and remains strong across all horizons, whereas histogram-based \ac{gb} benefits markedly from longer windows and \ac{knn} improves more gradually. The linear \ac{ridge} baseline remains ineffective throughout. Under the present benchmark and sensing assumptions, extending the horizon beyond 20\,ms therefore provides little additional benefit for the strongest classifier, while weaker nonlinear models still benefit from additional temporal context.

For \ac{fl}, the timing dependence in Table~\ref{tab:fl_timing_sensitivity} is much weaker. The best \ac{mae} decreases only modestly from 10.64 $\pm$ 0.36\,\% at 10\,ms to 9.92 $\pm$ 0.27\,\% at 50\,ms, and the model ranking remains stable across horizons. Histogram-based \ac{gb} and \ac{ridge} change little, while \ac{knn} improves only moderately. Longer windows therefore provide some benefit, but they do not remove the substantial localization error observed in this benchmark setting. Taken together, the timing analysis shows that decision horizon is part of the effective task definition rather than a secondary implementation choice: under the present evaluation setting, \ac{fc} reaches very high performance at short horizons, whereas localization retains physically material error across the same timing range.

\begin{table*}[pos=t]
\caption{Controlled observability sensitivity for fault classification. Macro-\ac{f1} under full, relay-pair, and single-relay sensing for \ac{mlp} and histogram-based \ac{gb} models at 20\,ms and 50\,ms. Full-observability values are the mean $\pm$ standard deviation over the five episode-grouped folds; the relay-pair and single-relay values are the mean over the four same-line relay pairs and the eight individual relays, respectively, with the min--max range across those configurations in brackets. Higher is better.}
\small
\label{tab:fc_observability}
\centering
\begin{tabular}{lcccc}
\toprule
\bfseries Observability & \bfseries \ac{mlp} (20\,ms) & \bfseries \ac{gb} (20\,ms) & \bfseries \ac{mlp} (50\,ms) & \bfseries \ac{gb} (50\,ms) \\
\midrule
Full         & \textbf{0.991 $\pm$ 0.001} & 0.745 $\pm$ 0.017 & \textbf{0.990 $\pm$ 0.005} & \textbf{0.982 $\pm$ 0.001} \\
Relay pair   & 0.989 (0.988--0.990) & \textbf{0.750} (0.717--0.777) & 0.987 (0.987--0.989) & 0.971 (0.968--0.975) \\
Single relay & 0.986 (0.975--0.993) & 0.734 (0.629--0.811) & 0.986 (0.966--0.991) & 0.952 (0.900--0.973) \\
\bottomrule
\end{tabular}
\end{table*}

\begin{table*}[pos=t]
\caption{Controlled observability sensitivity for fault localization. \ac{mae} in percent of normalized line length under full, relay-pair, and single-relay sensing for \ac{mlp} and histogram-based \ac{gb} models at 20\,ms and 50\,ms. Full-observability values are the mean $\pm$ standard deviation over the five episode-grouped folds; the relay-pair and single-relay values are the mean over the four same-line relay pairs and the eight individual relays, respectively, with the min--max range across those configurations in brackets. Lower is better.}
\small
\label{tab:fl_observability}
\centering
\begin{tabular}{lcccc}
\toprule
\bfseries Observability & \bfseries \ac{mlp} (20\,ms) & \bfseries \ac{gb} (20\,ms) & \bfseries \ac{mlp} (50\,ms) & \bfseries \ac{gb} (50\,ms) \\
\midrule
Full         & \textbf{10.20 $\pm$ 0.25} & \textbf{14.78 $\pm$ 0.25} & \textbf{9.92 $\pm$ 0.27} & \textbf{14.66 $\pm$ 0.20} \\
Relay pair   & 19.65 (17.40--21.67) & 21.43 (20.30--22.42) & 19.62 (17.23--21.63) & 21.37 (20.12--22.38) \\
Single relay & 22.22 (20.17--23.69) & 23.20 (22.46--23.94) & 22.17 (20.05--23.62) & 23.07 (22.19--23.77) \\
\bottomrule
\end{tabular}
\end{table*}

\subsection{Controlled Observability Sensitivity}
\label{sec:results_observability_sensitivity}

Tables~\ref{tab:fc_observability} and~\ref{tab:fl_observability} examine observability as a controlled framework dimension. Reduced-observability results are reported for the focused sensitivity panel, comprising the neural \ac{mlp} and histogram-based \ac{gb}, at the fixed 20\,ms reference horizon and at 50\,ms as a longer-window comparison. In all cases, the physical system, target construction, and validation protocol remain unchanged, so the observed differences can be attributed directly to information availability.

For \ac{fc}, the effect of reduced observability is limited. As shown in Table~\ref{tab:fc_observability}, the \ac{mlp} remains close to its full-observability baseline under both relay-pair and single-relay sensing, while histogram-based \ac{gb} shows only moderate degradation overall, aside from one small relay-pair improvement at 20\,ms. For \ac{fl}, the pattern is markedly different. Table~\ref{tab:fl_observability} shows substantial error increases under reduced observability for both models and at both horizons. Moving from full observability to relay-pair sensing already raises the \ac{mae} by roughly 6.7--9.7 percentage points, and single-relay sensing increases it further to roughly 8.4--12.3 points above the full-observability baseline. Taken together, these results show that, under the present benchmark and sensing assumptions, observability has only a limited effect on fault classification but is a dominant determinant of fault localization performance. In the language of the proposed framework, information availability is therefore part of the effective task definition rather than a secondary implementation detail.

\subsection{Conventional Protection Baselines Under a Shared Evaluation Structure}
\label{sec:results_baselines}
Table~\ref{tab:baselines_fl} places conventional impedance-based fault location alongside the learning models under the shared data partitions, targets, and task-specific evaluation structure. On clean, synchronized data the synchronized two-ended method is the strongest locator. Its per-window \ac{mae} is $5.42\%$ at 20\,ms, below the best classical-learning result in Table~\ref{tab:reference_20ms} ($10.20\%$), and, when a single estimate is latched per fault at settled current, it reaches $2.74\%$ at 20\,ms and $1.57\%$ at 50\,ms. This outcome is expected and useful for the framework: a physics-based estimator that optimally combines both synchronized terminals, and that additionally receives the true line parameters and the ground-truth loop selection denied to the reference learning models, should beat a general learner on clean data. The comparison is therefore not equal-information, and the conventional method's advantage is reported explicitly rather than presented as an equal-footing win. The single-ended reactance method is far weaker at 20\,ms ($41.9\%$) but improves sharply to $10.9\%$ at 50\,ms, because its reactance estimate requires a settled post-fault phasor that only the longer window provides. The symmetrical-component phase selector attains a fault-only macro-\ac{f1} of $0.83$ at 20\,ms and $0.84$ at 50\,ms. These values provide a descriptive threshold-based reference, but they are not compared numerically with the 11-class learning-model macro-\ac{f1} because the latter additionally includes the non-onset class and is evaluated over both onset-containing and non-onset windows.

Two framework points follow directly. First, the observability axis is exactly the axis that separates the conventional locators: two-ended location uses a relay pair, and single-ended location uses a single relay. Table~\ref{tab:baselines_obs} shows that under matched observability, though with the parameter advantage noted above, the conventional two-ended method ($2.74\%$ at relay-pair sensing) far outperforms the learning models restricted to the same relay pair ($19.6$--$21.4\%$), while single-ended location only becomes competitive with single-relay learning at the longer horizon. Second, the decision-horizon axis constrains the conventional methods too: single-ended location improves markedly, while the phase selector improves slightly, from 20\,ms to 50\,ms as the fundamental-cycle phasor settles, and the 10\,ms horizon is phasor-invalid. A further framework observation is that the onset-conditioned sample-validity rule, designed for and validated on the learning models, does not transfer unchanged to single-ended impedance location, which needs a settled phasor; the per-window and per-episode-settled columns of Table~\ref{tab:baselines_fl} quantify this gap. Sample validity is therefore method-class-dependent, a distinction the framework makes explicit rather than conceals.

\begin{table*}[pos=t]
\caption{\ac{fl} performance of learning and conventional impedance-based methods under shared episode partitions and task-specific metrics. \ac{mae} [\% line length], mean $\pm$ standard deviation over five folds. Per-window conventional results use the onset-conditioned evaluation windows, whereas settled results use a separately reported per-episode post-onset validity rule. Conventional locators additionally receive per-episode line parameters and ground-truth loop selection.}
\label{tab:baselines_fl}
\small
\centering
\begin{tabular}{llcc}
\toprule
\textbf{Method} & \textbf{Observability} & \textbf{20\,ms} & \textbf{50\,ms} \\
\midrule
\ac{mlp}        & full        & 10.20 $\pm$ 0.25 & $\phantom{0}$9.92 $\pm$ 0.27 \\
\ac{gb}        & full        & 14.78 $\pm$ 0.25 & 14.66 $\pm$ 0.20 \\
\ac{knn}        & full        & 20.10 $\pm$ 0.16 & 19.12 $\pm$ 0.14 \\
Ridge        & full        & 26.12 $\pm$ 0.36 & 26.13 $\pm$ 0.31 \\
\midrule
Two-ended (per window) & relay pair   & $\phantom{0}$5.42 $\pm$ 0.09 & $\phantom{0}$3.62 $\pm$ 0.23 \\
Two-ended (settled)    & relay pair   & $\phantom{0}$\textbf{2.74 $\pm$ 0.05} & $\phantom{0}$\textbf{1.57 $\pm$ 0.03} \\
One-ended (per window) & single relay & 156.2 $\pm$ 4.9 & 77.1 $\pm$ 5.6 \\
One-ended (settled)    & single relay & 41.9 $\pm$ 1.2 & 10.9 $\pm$ 0.6 \\
\bottomrule
\end{tabular}
\end{table*}

\subsection{Measurement-Fidelity Degradation}
\label{sec:results_fidelity}
Table~\ref{tab:fidelity_summary} summarizes measurement-fidelity robustness for the focused \ac{mlp}--\ac{gb} sensitivity panel at the 20\,ms horizon, reporting the clean value and the worst-case level across the three axes; the full per-level tables are given in Appendix~\ref{app:fidelity} (Tables~\ref{tab:fidelity_fl} and~\ref{tab:fidelity_fc}). The clean columns remain consistent with the corresponding reference results, confirming that the perturbation harness leaves the reference configuration effectively unchanged. Two findings stand out. First, degradation is monotonic within each evaluated axis, but the most damaging non-ideality is task- and model-dependent. Current-transformer saturation produces the largest localization degradation, whereas severe additive noise is particularly damaging for histogram-based \ac{gb} classification. Second, and more consequentially for the framework, robustness does not track clean predictive performance, and the robustness ranking of the models is itself task-dependent. For localization, the clean-best model is the most fragile: the \ac{mlp}, which attains the lowest clean error, inflates by approximately $1.2$ to $2.4\times$ across the severe perturbation levels at the evaluated 20\,ms horizon, whereas histogram-based \ac{gb} degrades far more gracefully ($1.0$ to $1.4\times$). For classification the ranking reverses (Table~\ref{tab:fidelity_fc}): the \ac{mlp} retains a macro-\ac{f1} above $0.95$ under every perturbation, while histogram-based \ac{gb} collapses under additive noise, falling from $0.74$ to $0.27$ at 10\,dB. Aggregate clean performance is therefore insufficient to rank models for deployment; robustness must be measured rather than inferred from nominal performance or model family. Robustness is a distinct evaluation dimension, and the framework makes it comparable across the evaluated learning models and tasks. Applying the same perturbation protocol to the conventional methods remains necessary for a direct cross-method-class robustness comparison.

\begin{table}[pos=t]
\caption{Compact measurement-fidelity summary at 20\,ms for the focused sensitivity panel: clean value and worst-case level across the three degradation axes (additive noise, current-transformer saturation, synchronization jitter). Full per-level results in Appendix~\ref{app:fidelity} (Tables~\ref{tab:fidelity_fl} and~\ref{tab:fidelity_fc}).}
\label{tab:fidelity_summary}
\small
\centering
\begin{tabular}{llcc}
\toprule
\textbf{Task (metric)} & \textbf{Model} & \textbf{Clean} & \textbf{Worst case} \\
\midrule
\ac{fl} (\ac{mae} [\%], $\downarrow$) & \ac{mlp} & 10.20 & 24.36 \\
\ac{fl} (\ac{mae} [\%], $\downarrow$) & \ac{gb}  & 14.78 & 20.56 \\
\midrule
\ac{fc} (macro-\ac{f1}, $\uparrow$)   & \ac{mlp} & 0.991 & 0.959 \\
\ac{fc} (macro-\ac{f1}, $\uparrow$)   & \ac{gb}  & 0.745 & 0.274 \\
\bottomrule
\end{tabular}
\end{table}

\begin{figure*}[pos=t]
  \centering
  \includegraphics[width=0.65\linewidth]{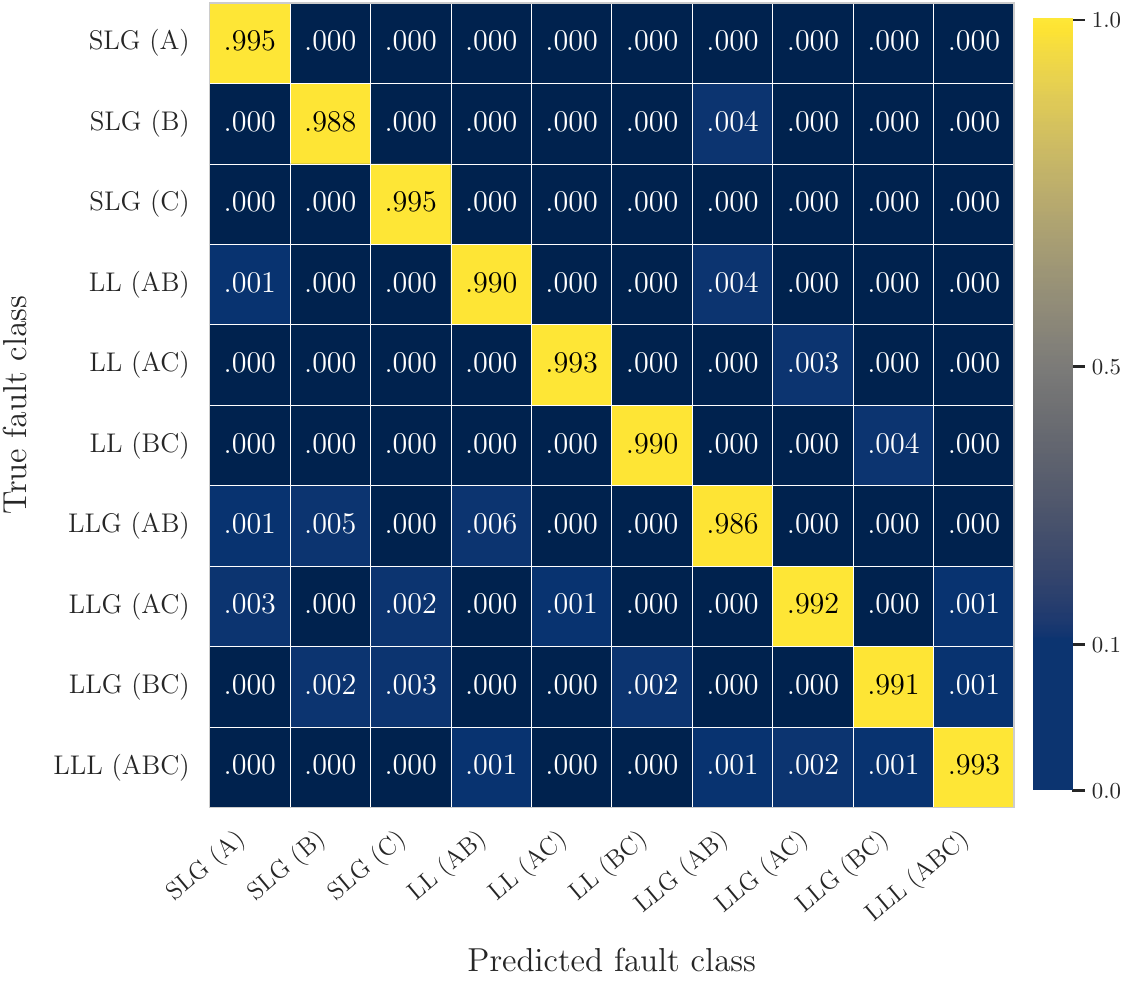}
  \caption{Row-normalized confusion matrix for \ac{fc} on fault-only cases under the 20\,ms \ac{mlp} configuration. Rows denote the true fault class and columns the predicted fault class. Diagonal entries indicate per-class recall, while off-diagonal cells represent misclassifications.}
  \label{fig:cm_fc_fault_only}
\end{figure*}

\begin{figure*}[pos=t]
    \centering
    \includegraphics[width=0.7\linewidth]{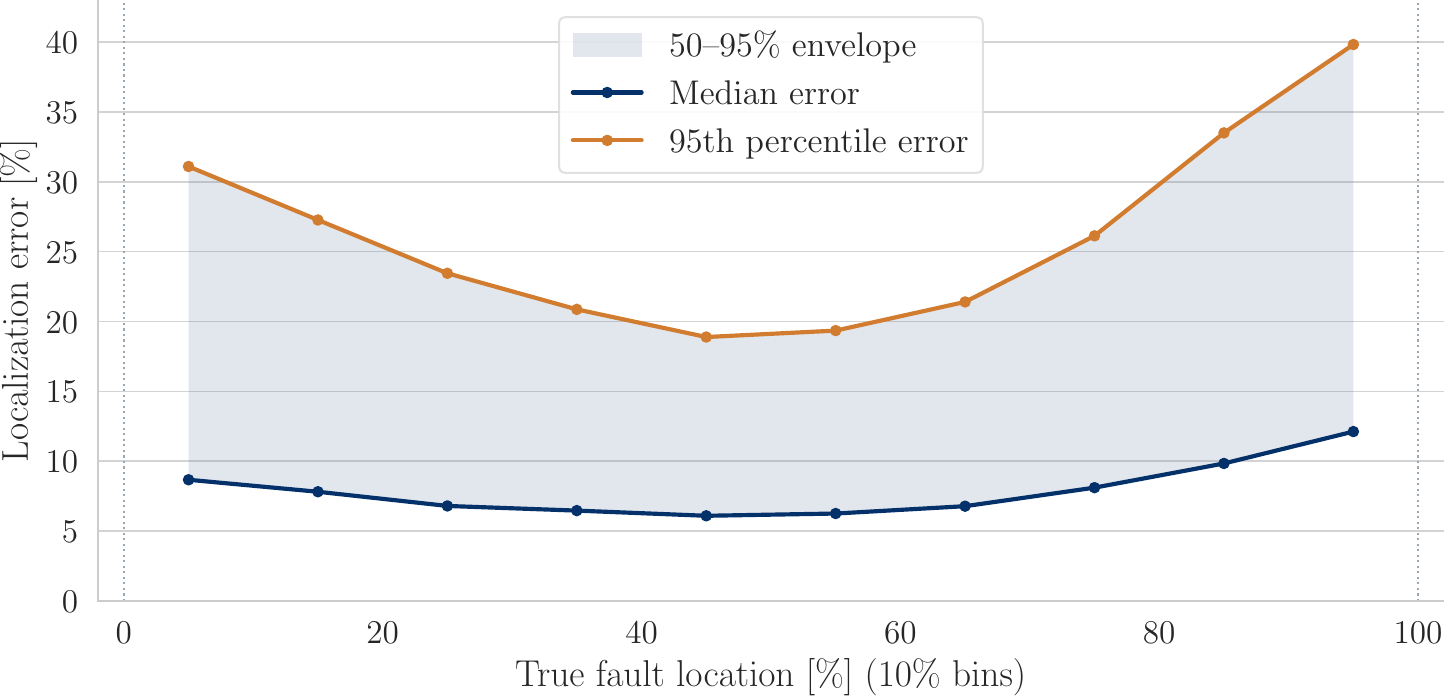}
    \caption{\Acl{fl} error as a function of the true fault position for the reference 20\,ms \ac{mlp} configuration (out-of-fold evaluation). Results are aggregated in 10\% line-length bins; the solid curve shows the median absolute error, and the shaded band spans the median to the 95th percentile.}
    \label{fig:fault_localization_error}
\end{figure*}

\subsection{Structured Diagnostics}
\label{sec:results_diagnostics}

Aggregate metrics summarize overall performance but do not show where models fail. The task-specific diagnostics therefore expose the structure behind the reported scores for the 20\,ms \ac{mlp} configuration.

For \ac{fc}, Fig.~\ref{fig:cm_fc_fault_only} shows that the errors are sparse and structured for the 20\,ms \ac{mlp}. The diagonal remains close to one across all fault classes, indicating that the strong aggregate macro-\ac{f1} is not driven by only a small subset of easy cases. Per-class fault-only \ac{f1} values remain uniformly high, ranging from 0.987 to 0.996. Because Fig.~\ref{fig:cm_fc_fault_only} is restricted to onset-containing fault cases, complementary pre-fault behavior is reported separately. On the held-out folds, the 20\,ms \ac{mlp} assigns a fault-type class to 3 of 117\,286 genuine pre-fault windows, corresponding to a pooled pre-fault-window false-positive rate of 0.0026\,\%. This is a bounded window-level diagnostic on the pre-fault portions of fault episodes. It should not be interpreted as an episode-level false-trip probability because no relay pickup, persistence, latching, or trip logic is modeled and the benchmark contains no dedicated non-fault disturbances. For \ac{fl}, Fig.~\ref{fig:fault_localization_error} shows that the residual error is not spatially uniform: median error is lowest in the mid-line region and increases toward the line ends, while the 95th-percentile envelope widens substantially near the boundaries. A coarse line-level summary shows the same tendency, with lower \ac{mae} for Line~1--2 A/B (9.08 and 9.18) than for Line~2--3 A/B (10.58 and 11.05). Taken together, these diagnostics indicate that \ac{fc} remains uniformly strong across classes, whereas \ac{fl} is shaped by topology- and location-dependent difficulty even in the strongest evaluated configuration.

\begin{table*}[pos=t]
\caption{Compact runtime summary for fault classification under the reference assumptions. Macro-\ac{f1} is shown for context.}
\label{tab:fc_runtime_compact}
\centering
\small

\begin{tabular}{lccccc}
\toprule
\textbf{Model} & \textbf{Window} & \textbf{\acs{f1}} & \textbf{Train [s]} & \textbf{Infer. [$\mu$s]} & \textbf{Thr. [k/s]} \\
\midrule
\ac{mlp} & 20\,ms & \textbf{0.991} & 603.7 & \textbf{7.32} & 136.7 \\
\ac{gb} & 20\,ms & 0.745 & 132.6 & 24.47 & 41.1 \\
\midrule
\ac{mlp} & 50\,ms & \textbf{0.990} & 1289.5 & \textbf{16.37} & 61.1 \\
\ac{gb} & 50\,ms & 0.982 & 1968.8 & 82.91 & 12.1 \\
\bottomrule
\end{tabular}
\end{table*}

\begin{table*}[pos=t]
\caption{Compact runtime summary for fault localization under the reference assumptions. \ac{mae} [\%] is shown for context.}
\label{tab:fl_runtime_compact}
\centering
\small
\begin{tabular}{lccccc}
\toprule
\textbf{Model} & \textbf{Window} & \textbf{\acs{mae}} & \textbf{Train [s]} & \textbf{Infer. [$\mu$s]} & \textbf{Thr. [k/s]} \\
\midrule
\ac{mlp} & 20\,ms & \textbf{10.20} & 346.3 & \textbf{6.45} & 155.0 \\
\ac{gb} & 20\,ms & 14.78 & 27.2 & 22.49 & 44.5 \\
\midrule
\ac{mlp} & 50\,ms & \textbf{9.92} & 2258.0 & \textbf{15.91} & 62.9 \\
\ac{gb} & 50\,ms & 14.66 & 140.4 & 49.36 & 20.3 \\
\bottomrule
\end{tabular}
\end{table*}

\subsection{Runtime Under the Reference Assumptions}
\label{sec:results_runtime}

Tables~\ref{tab:fc_runtime_compact} and~\ref{tab:fl_runtime_compact} summarize training time and model-side inference time under the reference assumptions. Runtime is measured on pre-windowed inputs. The reported inference times include feature scaling and model prediction, but exclude window extraction, disk I/O, communication overhead, and synchronization overhead. The values should therefore be interpreted as relative computational indicators under a common protocol, not as end-to-end relay latencies. For context, real-time processing at the 5\,ms stride requires approximately 200 windows per second per episode; all reported throughput figures exceed this threshold by more than one order of magnitude.

At the 20\,ms reference horizon, the \ac{mlp} combines the stronger predictive result with the lower inference time within the profiled \ac{mlp}--\ac{gb} subset, reaching a macro-\ac{f1} of 0.991 at 7.32\,$\mu$s per window for \ac{fc} and an \ac{mae} of 10.20 at 6.45\,$\mu$s for \ac{fl}. At 50\,ms, inference cost increases for both model families, but substantially more for histogram-based \ac{gb}, where \ac{fc} inference rises from 24.47 to 82.91\,$\mu$s per window. Within the profiled \ac{mlp}--\ac{gb} subset, the \ac{mlp} is therefore the more favorable option on both predictive and computational grounds.

\subsection{Stride Sensitivity Analysis}
\label{sec:results_stride}

A compact stride-sensitivity check for the representative \ac{mlp} and histogram-based \ac{gb} models (Appendix~\ref{app:stride}, Tables~\ref{tab:fc_stride_sensitivity} and~\ref{tab:fl_stride_sensitivity}) confirms that this secondary windowing choice does not overturn the task-level interpretation: stride can shift absolute scores and some model-family gaps -- notably, histogram-based \ac{gb} improves by more than $0.22$ macro-\ac{f1} at 20\,ms with larger strides -- but classification continues to approach its metric ceiling while localization retains material error, and the dominant effects remain task definition, timing, and observability.

\subsection{Hyperparameter Analysis}
\label{sec:results_robustness}

Single-parameter hyperparameter ablations for the representative \ac{mlp} and histogram-based \ac{gb} models (Appendix~\ref{app:robustness}, Table~\ref{tab:ablation_robustness_summary_main}) are diagnostic sensitivity checks rather than a hyperparameter search. The \ac{mlp} is highly stable for \ac{fc} while histogram-based \ac{gb} is more configuration-sensitive at the 20\,ms horizon, and both vary more for \ac{fl}; favorable settings improve absolute error but do not change the qualitative interpretation -- classification continues to approach its metric ceiling while localization retains material error, and the dominant effects remain task definition, decision horizon, and observability.

\subsection{Cross-Family Learned Baselines}
\label{sec:results_transformer}

To examine whether the task-level conclusions persist across model families, the comparison includes a classical \ac{mlp} and a compact \ac{cnn1d} trained from scratch. All models use the same episode-grouped five-fold protocol, sample-validity rules, decision horizons, and task-specific metrics, while retaining the model-specific input representation and preprocessing described in Section~\ref{sec:case_study_validation_models}. A pre-trained time-series foundation-model baseline (MOMENT-1-large, a frozen transformer with approximately 341 million parameters) is reported separately in Appendix~\ref{app:transformer}. Table~\ref{tab:transformer_summary} summarizes the main-text results.

At the 50\,ms horizon, the \ac{fl} \ac{mae} is $9.92\%$ for the \ac{mlp} and $9.09\%$ for the \ac{cnn1d}, while the pre-trained foundation-model baseline reaches $8.90\%$ (Appendix~\ref{app:transformer}). For \ac{fc}, the \ac{cnn1d} achieves the highest macro-\ac{f1} score at $0.999$, followed by the \ac{mlp} at $0.990$, with the foundation-model baseline at $0.989$. Despite the differences in absolute performance, the broader interpretation remains unchanged: under the shared evaluation setting, classification approaches its metric ceiling while localization retains material error across classical \ac{ml}, task-specific deep learning, and transfer from a pre-trained foundation model.

\begin{table*}[pos=t]
\caption{Main-text learned-baseline comparison under the shared episode-grouped five-fold protocol. \Ac{fc} is evaluated using macro-\ac{f1} ($\uparrow$), and \ac{fl} using \ac{mae} in percent of line length ($\downarrow$). Values are mean $\pm$ standard deviation across the five held-out folds; the best mean in each column is shown in bold. The pre-trained MOMENT-1-large results are reported separately in Appendix~\ref{app:transformer} (Table~\ref{tab:transformer_appendix}).}
\label{tab:transformer_summary}
\small
\centering
\begin{tabular}{lcccc}
\toprule
\textbf{Model} &
\textbf{\ac{fc} 20\,ms} &
\textbf{\ac{fc} 50\,ms} &
\textbf{\ac{fl} 20\,ms} &
\textbf{\ac{fl} 50\,ms} \\
\midrule
\ac{mlp} (reference) &
0.991 $\pm$ 0.001 &
0.990 $\pm$ 0.005 &
10.20 $\pm$ 0.25 &
\phantom{0}9.92 $\pm$ 0.27 \\

\ac{cnn1d} (from scratch) &
\textbf{0.999 $\pm$ 0.001} &
\textbf{0.999 $\pm$ 0.001} &
\textbf{\phantom{0}9.65 $\pm$ 0.67} &
\textbf{\phantom{0}9.09 $\pm$ 0.28} \\
\bottomrule
\end{tabular}
\end{table*}

\subsection{Framework-Level Synthesis}
\label{sec:results_synthesis}

Across the shared evaluation configuration and the controlled analyses, seven findings emerge.
First, \ac{fc} and \ac{fl} exhibit fundamentally different behavior despite identical sensing, timing, and validation assumptions.
Second, decision horizon and observability materially affect the apparent difficulty of both tasks and must therefore be treated as explicit evaluation dimensions.
Third, the classification--localization asymmetry persists across classical \ac{ml}, task-specific deep learning, and foundation-model transfer, indicating that it is not specific to a single model family.
Fourth, the synchronized two-ended conventional locator outperforms the learned locators on clean data when measurements from both terminals, line parameters, and ground-truth loop selection are available, demonstrating that information-set differences must be reported explicitly.
Fifth, measurement degradation shows that clean-data performance does not determine robustness and that robustness rankings can differ between tasks.
Sixth, the directional fault-resistance holdout reveals model- and direction-dependent performance differences (Appendix~\ref{app:generalization}).
Seventh, stride and hyperparameter choices can affect differences between model families, while structured diagnostics and runtime measurements reveal limitations that aggregate metrics alone do not expose.
Together, these findings support the central premise of the framework: performance in machine-learning-based protection can be interpreted only relative to the complete evaluation setting.

\section{Discussion}
\label{sec:discussion}

This paper proposes a standardized evaluation framework for \ac{ml}-based power system protection and demonstrates it through one bounded case study. The main contribution is not a claim about one universally best model or one definitive benchmark result, but a reusable way to specify, evaluate, and interpret protection studies before model performance is compared. The framework defines what must be fixed and reported; the case study shows what becomes visible when those requirements are met.

\subsection{Main Contributions of the Framework}
\label{sec:discussion_framework}

The reusable contribution is the seven-step evaluation logic and the associated reporting structure. Its central claim is that reported results are only scientifically interpretable when the effective inference problem -- task, observability, timing, and validation -- is made explicit before model performance is compared. Under this perspective, the value of the case study is not that it adds another benchmark leaderboard, but that it shows what becomes visible once these dimensions are fixed under a common protocol: \ac{fc} and \ac{fl} behave differently even under shared assumptions, and observability materially changes the apparent difficulty of the task.

The sensitivity analyses further show that even model-family comparisons are not invariant to seemingly secondary configuration choices. Histogram-based \ac{gb} improves substantially for \ac{fc} at the 20\,ms horizon under alternative stride and hyperparameter settings, while the \ac{mlp} remains comparatively stable. Thus, part of the observed \ac{mlp}--\ac{gb} gap in the default configuration reflects configuration sensitivity rather than only model-family capability. This illustrates why window construction, preprocessing, model configuration, and validation protocol must be reported as part of the evaluated protection problem, rather than treated as incidental implementation details.

The framework also accommodates multi-function protection schemes in which \ac{fd}, \ac{fc}, and \ac{fl} are required jointly rather than in isolation. In such settings, Step~1 must enumerate each protection function and its learning formulation explicitly, while Steps~4 and~5 must state whether timing constraints and target definitions are shared or task-specific. This avoids the interpretive ambiguity that arises when distinct protection objectives share a single aggregate metric, masking task-specific failure modes.

The framework's novelty therefore lies in integrating complementary protection and machine-learning requirements into one operational evaluation workflow, not in claiming that its individual dimensions are unprecedented. Applied retrospectively, the reporting package also exposes which assumptions would need to be reconstructed from supplementary material or clarified by the original authors.

The exact macro-\ac{f1} values, \ac{mae} values, model rankings, and the persistent localization error are specific to PROTECT-90, the Double Line topology, the chosen sensing regimes, the representative model set, and the stated configuration choices used here. These findings should therefore be read as benchmark-conditioned evidence, not as universal statements about all protection settings. What the community can reuse is not the specific performance profile of this benchmark, but the evaluation structure, reporting package, and interpretation logic that tie empirical results to explicit protection assumptions.

\subsection{Cross-Method-Class Comparison, Robustness, and Consistency}
\label{sec:discussion_synthesis}
The conventional comparisons show that the framework applies beyond learning models, because their performance likewise depends on observability, decision horizon, auxiliary inputs, and sample-validity rules. The observability axis is the axis that separates one- from two-terminal fault location, and under matched observability the synchronized two-ended locator outperforms the learning models on clean data; the decision-horizon axis is the budget that governs phasor estimability, and single-ended location and the phase selector both improve as the fundamental cycle settles. Recent work comparing dynamic-state-estimation-based protection with Transformer-based diagnosis illustrates a complementary division of functions: the physics-based method provides rapid anomaly detection, whereas the Transformer differentiates physical faults from measurement attacks and supplies measurement-level diagnostic information~\cite{abukhousa_transformer_2026}. Such combinations should therefore be evaluated using function-specific timing budgets, targets, and outputs rather than through a single aggregate performance ranking.

Within the \ac{mlp}--\ac{gb} fidelity comparison, robustness does not track clean predictive performance and the ranking is task-dependent: for localization, the lower-error \ac{mlp} degrades more strongly than \ac{gb}, whereas for classification the ranking reverses. Aggregate clean performance therefore does not by itself rank models for deployment. More broadly, realism and robustness, together with cross-method-class comparability, are themselves evaluation dimensions that the framework exposes.

These results are consistent with the surrounding literature on the same benchmark. The reference and timing errors agree with the authors' controlled model-comparison study on this benchmark~\cite{oelhaf_controlled_2026}, which likewise fixes default hyperparameters; the \ac{fc} and \ac{fl} reference and timing values reported here are shared with that controlled-comparison study, and this overlap is disclosed explicitly, as it is for the companion robustness study. The sensor-availability robustness that the present study deliberately does not repeat is characterized in a companion paper on the same data~\cite{oelhaf_robustness_2026}; the lower localization error reported there under performance-tuned hyperparameters is consistent with the present study's diagnostic ablation, which shows the favorable-hyperparameter error moving toward that value. This is consistent with the framework rather than an inconsistency, and it underscores the requirement to report the hyperparameter procedure.

Finally, the framework fixes and reports the effective inference problem so that comparisons are legible; it does not by itself guarantee that a benchmark is representative of the physical-operational problem, and representativeness and critical-condition coverage remain open problems that the framework highlights but does not solve. A utility or relay manufacturer could use the framework as a study-design and reporting checklist for internal and third-party evaluations, not as a deployment certification. Operationally, adoption can proceed in three stages. First, the protection task, physical scope, information access, and timing budget are frozen before model development. Second, nominal, degraded, and critical-condition test matrices are defined together with task-specific acceptance criteria. Third, a versioned reporting package records data provenance, preprocessing boundaries, validation groups, model configuration, diagnostics, and runtime conditions. This supports auditable internal studies, supplier comparisons, and procurement-oriented evaluation without replacing formal product qualification or protection certification.

\subsection{Comparability, Reproducibility, and Boundaries}
\label{sec:discussion_comparability_boundaries}

The framework improves comparability by shifting the scientific point of comparison from models alone to the evaluation setting that defines the problem those models are solving. In much of the existing literature, grid assumptions, sensing access, decision windows, target definitions, and split protocols vary simultaneously, so similar headline scores may correspond to different protection problems. Here, these dimensions are fixed and reported explicitly before model performance is interpreted, making it easier to distinguish task-dependent conclusions from differences induced by information availability, window construction, model configuration, or validation choices. The same logic improves reproducibility: reproducibility is not only a matter of releasing code, but of making the evaluation path inspectable enough that another group can reconstruct, critique, and reuse it under the same stated conditions. Moving from reproducible research evaluation toward deployment requires additional verification and validation procedures for neural-network-based protection schemes~\cite{mederer_verification_2025}.

Openness of the underlying data strengthens reproducibility, but full data release is not always feasible: protection recordings and grid models are frequently subject to confidentiality, cybersecurity, and intellectual-property constraints at utilities and equipment vendors. The framework remains applicable under these limits, because it preserves comparability at the level of study design. Even when raw waveforms cannot be shared, a complete reporting package -- the declared physical scope, observability, timing, target and sample-validity definitions, and validation protocol -- allows a result to be interpreted, critiqued, and reused under its stated conditions. Data openness and its institutional constraints are thus reported as an explicit study dimension rather than treated as an all-or-nothing requirement.

That said, the present instantiation is bounded in several respects. Empirically, it covers one public \ac{emt} benchmark derived from a single 90\,kV Double Line topology. The observed task-dependent performance asymmetry between \ac{fc} and \ac{fl} may not generalize to meshed or ring-bus topologies, where overlapping current paths can make fault type discrimination substantially harder while simultaneously providing richer spatial information for localization. In terms of realism, the reference setting assumes synchronized, noise-free, continuously available measurements. Measurement-fidelity non-idealities (additive noise, current-transformer saturation, and synchronization jitter) are characterized as controlled sensitivity axes in Section~\ref{sec:results_fidelity}, and sensor-availability degradation on the same data is characterized in the companion study~\cite{oelhaf_robustness_2026}; communication latency and broader non-fault disturbances remain outside the present scope. In particular, security-critical non-fault events -- such as transformer inrush, capacitor-bank switching, motor starting, and power swings -- are not represented in the present benchmark. The pre-fault-window false-positive rate reported among the structured diagnostics is therefore a bounded diagnostic rather than an episode-level false-trip probability; a systematic false-trip and failure-to-trip security assessment under such conditions is left to future work. The model coverage is investigation-specific: the broad classical panel is used for the reference and timing analyses, the \ac{mlp}--\ac{gb} subset for focused sensitivity analyses, and the \ac{mlp}--\ac{cnn1d}--MOMENT panel for the cross-family comparison; conventional protection methods are evaluated separately in Section~\ref{sec:results_baselines}. Finally, the reported runtime values are model-side indicators on pre-windowed inputs, not end-to-end relay latencies; the \ac{cnn1d} and MOMENT baselines were not profiled for runtime or memory and therefore do not support conclusions regarding deployment suitability.

The framework applies equally to field-recording settings, where ground-truth metadata is unavailable or uncertain. In such cases, Step~2 must flag the data source and its label provenance explicitly; Step~5 must document the labeling procedure and its associated uncertainty; and Step~4 must adapt the temporal reference to a proxy such as relay trip time if fault inception is unknown. Label uncertainty then becomes a declared study dimension that propagates into the interpretation of all downstream metrics, rather than a hidden assumption.

\subsection{Future Extensions of the Framework}
\label{sec:discussion_future}

The next step is not simply to add more models, but to extend the same evaluation logic to broader and more realistic protection settings. This includes additional topologies and operating regimes, broader conventional protection methods, and additional recurrent, topology-aware, and end-to-end adapted foundation-model baselines under the same protocol, including graph-based protection models~\cite{kordowich_graph_2025}. It also includes extension of the disturbance space toward higher-resistance and high-impedance fault regimes beyond the present benchmark range, as well as security-critical non-fault events. Sensing studies should extend the present measurement-fidelity analyses toward physically detailed \ac{ct}/\ac{vt} models, correlated or nonstationary noise, missing channels, and communication latency, jitter, or loss. In this context, feature- and channel-selection methods provide one route for studying which measurements are most relevant under reduced or degraded observability~\cite{kordowich_feature_2026}.

For studies working with smaller datasets, the grouped cross-validation protocol requires careful adaptation. When the number of independent groups per fold is low, fold-level estimates become sensitive to partition randomness. In such settings, leave-one-group-out cross-validation is preferable; results should additionally verify class coverage across folds before aggregate metrics are interpreted. At the reporting level, future work should complement aggregate predictive metrics with outputs more directly tied to protection security, such as false-positive behavior under extreme non-fault conditions, and move toward end-to-end latency accounting where deployment relevance is the goal. The feasible extension of the framework will depend on the availability of suitable public datasets and on access to information that is often vendor-locked, especially with respect to hardware behavior, instrument nonidealities, and relay implementation details. Pursued systematically, these extensions would move the framework from a controlled research standard toward an evaluation basis for deployment-relevant protection assessment.

\section{Conclusion}
\label{sec:conclusion}
Reported performance in machine-learning-based power system protection is not a property of the model alone. It depends on the protection objective, physical system, available measurements, decision time, target construction, validation protocol, and reported evidence. The proposed seven-step framework makes these conditions explicit and treats evaluation design as part of the scientific contribution. The unit of comparison therefore becomes the complete inference problem, not an isolated algorithm or headline score.

The PROTECT-90 case study shows why this matters. Under a fixed physical scope, sensing configuration, timing definition, and episode-grouped validation protocol, onset-conditioned fault classification approached its metric ceiling while fault localization retained physically material error. Reduced observability had little effect on classification but approximately doubled localization error. The synchronized two-ended locator outperformed the waveform-based learners when given measurements from both terminals, line parameters, and ground-truth loop selection. Measurement degradation also showed that the best clean-data model need not be the most robust. These results are specific to the evaluated benchmark and do not establish universal rankings. Their value lies in showing which assumptions produce each result.

The framework is not yet a deployment certificate; it defines the evidence needed for future qualification and certification. Reaching that stage requires evaluation across additional topologies and operating regimes, broader conventional and learning-based method families, uncertain event timing, field and hardware-in-the-loop recordings, security-critical non-fault disturbances, instrument-transformer nonidealities, communication degradation, and end-to-end latency. Near-perfect scores on isolated benchmarks are insufficient. Protection models need evidence that remains interpretable and comparable as tasks, sensors, and operating conditions change. By making those conditions explicit and testable, the proposed framework turns separate studies into cumulative engineering evidence and provides a path toward dependable, auditable, and certifiable machine-learning-based protection.

\section*{Data and Code Availability}

The case study is based on the publicly available PROTECT-90 dataset. The accompanying paper documents the simulation design, waveform structure, metadata, and intended benchmark use~\cite{oelhaf_protect-90_2026}; dataset version 1.0.0 is available through Zenodo at \href{https://doi.org/10.5281/zenodo.21109169}{10.5281/zenodo.21109169}~\cite{kordowich_protect-90_2026}. The corresponding framework implementation, including code for preprocessing, windowing, task construction, leakage-aware validation, and result reporting, is available at \href{https://github.com/julianoelhaf/protection-eval-framework}{https://github.com/julianoelhaf/protection-eval-framework}.

\printcredits

\section*{Declaration of competing interest}
\noindent
The authors declare that they have no known competing financial interests or personal relationships that could have appeared to influence the work reported in this paper.

\section*{Acknowledgment}
\noindent
This project was funded by the Deutsche Forschungsgemeinschaft (DFG, German Research Foundation) - 535389056.

\newpage
\bibliography{references}

\appendix

\counterwithin{table}{section}
\counterwithin{figure}{section}
\counterwithin{equation}{section}

\section{Detailed Reduced-Observability Results}
\label{app:observability}

This appendix reports relay-level reduced-observability results for the representative 50\,ms \ac{mlp} \ac{fl} configuration. Table~\ref{tab:fl_observability_detail_mlp_50ms} complements the aggregate observability summary in Table~\ref{tab:fl_observability} by comparing single-terminal and same-line two-terminal sensing for each protected line. The results show that two-terminal same-line sensing consistently reduces localization error relative to either individual terminal, while the remaining variation across lines indicates that measurement usefulness is topology-dependent rather than uniform across the grid.

\begin{table*}[pos=ht]
\centering
\small
\caption{Relay-level \ac{fl} error for the 50\,ms \ac{mlp} regressor under reduced observability. The improvement is computed relative to the better of the two single-terminal settings for each line.}
\label{tab:fl_observability_detail_mlp_50ms}
\begin{tabular}{lcccc}
\toprule
\textbf{Line} &
\multicolumn{2}{c}{\textbf{Single-terminal \acs{mae} [\%]}} &
\textbf{Two-terminal} &
\textbf{Improvement} \\
\cmidrule(lr){2-3}
& \textbf{Terminal 1} & \textbf{Terminal 2} & \textbf{\acs{mae} [\%]} & \textbf{[\% points]} \\
\midrule
01--02A & 21.948 & 20.055 & 17.229 & 2.826 \\
01--02B & 22.316 & 22.712 & 20.814 & 1.502 \\
02--03A & 20.510 & 23.202 & 18.797 & 1.712 \\
02--03B & 22.997 & 23.620 & 21.630 & 1.367 \\
\bottomrule
\end{tabular}
\end{table*}

\Ac{fl} performance under observability matched between conventional and learning methods is reported in Table~\ref{tab:baselines_obs}; these values also appear, differently organized, in Table~\ref{tab:baselines_fl}.

\begin{table}[pos=ht]
\caption{\ac{fl} performance by observability at 20\,ms for sensing-matched conventional and learning methods. Conventional values use the separately reported per-episode settled estimates and additionally receive episode-specific line parameters and ground-truth loop selection; the comparison is therefore sensing-matched but not input- or sample-validity-matched. \ac{mae} [\% line length]; lower is better.}
\label{tab:baselines_obs}
\small
\centering
\begin{tabular}{llccc}
\toprule
\textbf{Observability} & \textbf{Conventional} & \textbf{Conv.} & \textbf{\ac{mlp}} & \textbf{\ac{gb}} \\
\midrule
Full         & (n/a)                & --   & 10.20 & 14.78 \\
Relay pair   & two-ended synchr.    & \textbf{2.74} & 19.65 & 21.43 \\
Single relay & one-ended reactance  & 41.9 & 22.22 & 23.20 \\
\bottomrule
\end{tabular}
\end{table}

\section{Detailed Measurement-Fidelity Results}
\label{app:fidelity}
This appendix reports the full per-level measurement-fidelity results summarized compactly in Table~\ref{tab:fidelity_summary}. Tables~\ref{tab:fidelity_fl} and~\ref{tab:fidelity_fc} give localization and classification degradation, respectively, across all evaluated levels of additive noise, current-transformer saturation, and synchronization jitter, for the focused \ac{mlp}--\ac{gb} sensitivity panel at the 20\,ms horizon.

\begin{table}[pos=ht]
\caption{\ac{fl} robustness to measurement degradation at 20\,ms. The clean row reproduces the five-fold reference statistic. For each perturbed level, the five realizations are first averaged within each fold; values are then reported as mean $\pm$ standard deviation across the five fold-level means. Lower is better.}
\label{tab:fidelity_fl}
\small
\centering
\begin{tabular}{llcc}
\toprule
\textbf{Axis} & \textbf{Level} & \textbf{\ac{mlp}} & \textbf{\ac{gb}} \\
\midrule
Clean         & --        & 10.20 $\pm$ 0.25 & 14.78 $\pm$ 0.25 \\
Noise         & 20\,dB    & 10.48 $\pm$ 0.26 & 14.91 $\pm$ 0.24 \\
Noise         & 10\,dB    & 12.55 $\pm$ 0.35 & 15.75 $\pm$ 0.22 \\
\ac{ct} saturation & $c=0.5$   & 15.91 $\pm$ 0.24 & 17.44 $\pm$ 0.35 \\
\ac{ct} saturation & $c=0.3$   & \textbf{24.36 $\pm$ 0.67} & 20.56 $\pm$ 0.37 \\
Jitter        & 2 samples & 12.30 $\pm$ 1.35 & 15.05 $\pm$ 0.26 \\
Jitter        & 4 samples & 15.06 $\pm$ 2.94 & 15.28 $\pm$ 0.31 \\
\bottomrule
\end{tabular}
\end{table}

\begin{table}[pos=ht]
\caption{\ac{fc} robustness to measurement degradation at 20\,ms. The clean row reproduces the five-fold reference statistic. For each perturbed level, the five realizations are first averaged within each fold; values are then reported as mean $\pm$ standard deviation across the five fold-level means. Higher is better.}
\label{tab:fidelity_fc}
\small
\centering
\begin{tabular}{llcc}
\toprule
\textbf{Axis} & \textbf{Level} & \textbf{\ac{mlp}} & \textbf{\ac{gb}} \\
\midrule
Clean         & --        & 0.991 $\pm$ 0.001 & 0.745 $\pm$ 0.017 \\
Noise         & 20\,dB    & 0.989 $\pm$ 0.003 & 0.592 $\pm$ 0.024 \\
Noise         & 10\,dB    & 0.982 $\pm$ 0.011 & \textbf{0.274 $\pm$ 0.028} \\
\ac{ct} saturation & $c=0.5$   & 0.982 $\pm$ 0.002 & 0.682 $\pm$ 0.021 \\
\ac{ct} saturation & $c=0.3$   & 0.959 $\pm$ 0.003 & 0.534 $\pm$ 0.021 \\
Jitter        & 4 samples & 0.983 $\pm$ 0.006 & 0.682 $\pm$ 0.034 \\
\bottomrule
\end{tabular}
\end{table}

\section{Directional Holdout Across Fault-Resistance Ranges}
\label{app:generalization}

The reference and preceding sensitivity analyses preserve the same fault-resistance distribution between training and test data. To examine performance under a directional fault-condition holdout, the episode set is partitioned into disjoint training and test ranges according to fault resistance $R_f$. Two directional splits are evaluated at the 20\,ms reference horizon: training on the lower 80\% of the $R_f$ range and testing on the upper 20\%, and training on the upper 80\% and testing on the lower 20\%. The split is performed by episode, so no windows from the same episode occur in both partitions. Table~\ref{tab:generalization_shift} reports the focused \ac{mlp}--\ac{gb} sensitivity panel under these directional holdouts. The ordinary episode-grouped five-fold result is included for context but is not a training-size-matched control.

\begin{table*}[pos=ht]
\caption{Directional fault-resistance holdout at 20\,ms. Models are trained on the lower 80\% and tested on the upper quintile of $R_f$, or vice versa. \ac{fc}: macro-\ac{f1}; \ac{fl}: \ac{mae} [\% line length]. The episode-grouped five-fold result is shown for context and is not a training-size-matched control. Shifted \ac{mlp} columns are mean\,$\pm$\,standard deviation across five explicitly varied model seeds (0--4); the histogram-based \ac{gb} columns report one run with seed 42.}
\label{tab:generalization_shift}
\small
\centering
\begin{tabular}{llccc}
\toprule
\textbf{Task} & \textbf{Model} & \textbf{Five-fold reference} & \textbf{High-$R_f$ test} & \textbf{Low-$R_f$ test} \\
\midrule
\ac{fc} (macro-\ac{f1}) & \ac{mlp} & 0.991 & $0.975\pm0.004$ & $0.979\pm0.007$ \\
\ac{fc} (macro-\ac{f1}) & \ac{gb}  & 0.745 & 0.731 & 0.725 \\
\midrule
\ac{fl} (\ac{mae} [\%]) & \ac{mlp} & 10.20 & $10.33\pm0.22$ & $12.34\pm0.63$ \\
\ac{fl} (\ac{mae} [\%]) & \ac{gb}  & 14.78 & 16.79 & 14.30 \\
\bottomrule
\end{tabular}
\end{table*}

Under these directional fault-resistance holdouts, the representative models retain broadly similar performance, although the comparison with the five-fold reference is descriptive rather than training-size matched. The \ac{mlp} classifier remains within $0.02$ macro-\ac{f1} of its baseline, while the \ac{mlp} locator changes only slightly on the high-resistance range ($10.33\pm0.22\%$, compared with the $10.20\%$ five-fold in-distribution reference) and degrades moderately to $12.34\pm0.63\%$ on the low-resistance range. Histogram-based \ac{gb} localization is more sensitive to the shift toward higher resistance, with the \ac{mae} increasing to $16.79\%$, whereas its error decreases slightly on the low-resistance test range. The observed differences are direction- and model-dependent, showing that fault-resistance coverage is a distinct robustness axis that should be reported explicitly. This within-benchmark experiment is illustrative rather than exhaustive; shifts in topology, loading, fault families, and data source remain future instantiations under the same evaluation framework.

\section{Class Distribution and Test-Set Sizes}
\label{app:class_distribution}

Table~\ref{tab:class_distribution} reports the class distribution of the 20\,ms onset-conditioned classification windows. The target is dominated by the non-onset class ($89.66\%$ of windows), while the rarest fault-type class accounts for $0.83\%$, an imbalance of roughly $108{:}1$; this motivates macro-averaged \ac{f1} rather than accuracy as the classification metric. Each grouped five-fold test partition contains ${\approx}52.3$k windows over $1{,}804$--$1{,}805$ episodes for \ac{fc} (all windows), and ${\approx}5.4$k fault-onset windows over the same episodes for \ac{fl} (fault-only).

\begin{table*}[pos=ht]
\centering
\small
\caption{Class distribution of the 20\,ms fault-classification windows (\texttt{event\_type}; $N = 261{,}638$ windows over $9{,}022$ episodes).}
\label{tab:class_distribution}
\begin{tabular}{llrr}
\toprule
\textbf{Class} & \textbf{Type} & \textbf{Windows} & \textbf{Share} \\
\midrule
Non-onset       & --       & 234\,572            & 89.66\% \\
AG / BG / CG    & 1ph--g   & 2295 / 2259 / 2283  & 0.88 / 0.86 / 0.87\% \\
AB / BC / CA    & 2ph      & 2175 / 2262 / 2316  & 0.83 / 0.86 / 0.89\% \\
ABG / BCG / CAG & 2ph--g   & 2175 / 2205 / 2277  & 0.83 / 0.84 / 0.87\% \\
ABC             & 3ph      & 6819                & 2.61\% \\
\midrule
Total           &          & 261\,638            & 100\% \\
\bottomrule
\end{tabular}
\end{table*}

\section{Pre-Trained Foundation-Model Baseline}
\label{app:transformer}
This appendix reports the pre-trained foundation-model baseline from the cross-family comparison in Section~\ref{sec:results_transformer}. MOMENT-1-large, a frozen time-series transformer with approximately 341 million parameters, is evaluated using either a linear probe or a two-layer \ac{mlp} head on its per-channel embeddings. The fixed input-adaptation, embedding-aggregation, readout, optimization, and stopping settings are documented in the released experiment configuration. Both readouts follow the same episode-grouped five-fold protocol, sample-validity rules, decision horizons, and task-specific metrics as the main-text learned models, with fold-local standardization and independent regeneration of training and held-out embeddings in every outer fold. Table~\ref{tab:transformer_appendix} reports their results alongside the reference \ac{mlp} and \ac{cnn1d}. The nonlinear head outperforms the linear probe on both tasks, indicating that the frozen embeddings contain task-relevant information that is not fully linearly accessible. At 50\,ms, the \ac{mlp} head achieves the lowest mean \ac{fl} error in the fixed cross-family comparison ($8.90\%$), while the main interpretation remains unchanged: classification approaches its metric ceiling, whereas localization retains material error across classical \ac{ml}, task-specific deep learning, and foundation-model transfer.

\begin{table*}[pos=ht]
\caption{Pre-trained foundation-model (MOMENT-1-large) baseline under the shared episode-grouped five-fold protocol, with the main-text \ac{mlp} and \ac{cnn1d} repeated for reference. \ac{fc}: macro-\ac{f1} ($\uparrow$); \ac{fl}: \ac{mae} [\% line length] ($\downarrow$). Mean $\pm$ standard deviation over 5 folds; best mean per column in bold.}
\label{tab:transformer_appendix}
\small
\centering
\begin{tabular}{lcccc}
\toprule
\textbf{Model} & \textbf{\ac{fc} 20\,ms} & \textbf{\ac{fc} 50\,ms} & \textbf{\ac{fl} 20\,ms} & \textbf{\ac{fl} 50\,ms} \\
\midrule
\ac{mlp} (reference)     & 0.991 $\pm$ 0.001 & 0.990 $\pm$ 0.005 & 10.20 $\pm$ 0.25 & \phantom{0}9.92 $\pm$ 0.27 \\
\ac{cnn1d} (from scratch)    & \textbf{0.999 $\pm$ 0.001} & \textbf{0.999 $\pm$ 0.001} & \phantom{0}\textbf{9.65 $\pm$ 0.67} & \phantom{0}9.09 $\pm$ 0.28 \\
MOMENT -- linear probe  & 0.953 $\pm$ 0.005 & 0.975 $\pm$ 0.002 & 14.82 $\pm$ 0.10 & 14.99 $\pm$ 0.09 \\
MOMENT -- \ac{mlp} head & 0.985 $\pm$ 0.002 & 0.989 $\pm$ 0.001 & 10.59 $\pm$ 0.22 & \phantom{0}\textbf{8.90 $\pm$ 0.12} \\
\bottomrule
\end{tabular}
\end{table*}

\section{Detailed Stride Sensitivity Analysis}
\label{app:stride}

Tables~\ref{tab:fc_stride_sensitivity} and~\ref{tab:fl_stride_sensitivity} report a compact stride-sensitivity check for the representative \ac{mlp} and histogram-based \ac{gb} models. The purpose is not stride optimization, but to assess whether a secondary windowing choice changes the interpretation of the reference evaluation setting. For \ac{fc}, the \ac{mlp} remains largely stable across the tested strides, whereas histogram-based \ac{gb} is more sensitive at short and intermediate windows. In particular, its 20\,ms performance improves by more than 0.22 macro-\ac{f1} at larger strides, showing that some apparent model differences can reflect configuration choices rather than only model-family capability. At 50\,ms, both models remain close to baseline. For \ac{fl}, the \ac{mlp} is stable at 10\,ms but degrades at larger strides for the 20\,ms and 50\,ms windows, while histogram-based \ac{gb} shows mixed changes with no consistent gain from increasing stride. Overall, stride can affect absolute scores and relative model gaps in selected settings, and the large improvement of histogram-based \ac{gb} at 20\,ms shows that model-family comparisons can be configuration-dependent. However, this sensitivity does not overturn the task-level interpretation of the case study: classification continues to approach its metric ceiling while localization retains material error, and the broader conclusions remain shaped primarily by task definition, timing, and observability.

\begin{table*}[pos=ht]
\caption{Compact stride sensitivity for fault classification. Baseline performance is reported at the default 5\,ms stride; additional columns show $\Delta$ macro-\ac{f1} relative to that baseline for larger strides. Positive values indicate improvement.}
\small
\label{tab:fc_stride_sensitivity}
\centering
\begin{tabular}{llcccc}
\toprule
\textbf{Model} & \textbf{Window} & \textbf{Baseline \acs{f1}} & \textbf{$\Delta$ @ 10\,ms} & \textbf{$\Delta$ @ 20\,ms} & \textbf{$\Delta$ @ 50\,ms} \\
\midrule
\ac{mlp} & 10\,ms & 0.985 & +0.003 & -- & -- \\
\ac{mlp} & 20\,ms & 0.991 & -0.008 & -0.010 & -- \\
\ac{mlp} & 50\,ms & 0.990 & -0.004 & -0.002 & -0.003 \\
\midrule
\ac{gb} & 10\,ms & 0.418 & +0.100 & -- & -- \\
\ac{gb} & 20\,ms & 0.745 & +0.221 & +0.225 & -- \\
\ac{gb} & 50\,ms & 0.982 & -0.005 & -0.003 & +0.004 \\
\bottomrule
\end{tabular}
\end{table*}

\begin{table*}[pos=ht]
\caption{Compact stride sensitivity for fault localization. Baseline performance is reported at the default 5\,ms stride; additional columns show $\Delta$\ac{mae} relative to that baseline for larger strides. Negative values indicate improvement.}
\small
\label{tab:fl_stride_sensitivity}
\centering
\begin{tabular}{llcccc}
\toprule
\textbf{Model} & \textbf{Window} & \textbf{Baseline \acs{mae}} & \textbf{$\Delta$ @ 10\,ms} & \textbf{$\Delta$ @ 20\,ms} & \textbf{$\Delta$ @ 50\,ms} \\
\midrule
\ac{mlp} & 10\,ms & 10.64 & +0.00 & -- & -- \\
\ac{mlp} & 20\,ms & 10.20 & +0.93 & +0.13 & -- \\
\ac{mlp} & 50\,ms & 9.92 & +1.10 & +1.24 & +1.44 \\
\midrule
\ac{gb} & 10\,ms & 14.65 & +0.00 & -- & -- \\
\ac{gb} & 20\,ms & 14.78 & +0.28 & -0.36 & -- \\
\ac{gb} & 50\,ms & 14.66 & +0.49 & +0.65 & -0.20 \\
\bottomrule
\end{tabular}
\end{table*}

\section{Detailed Hyperparameter Analysis Results}
\label{app:robustness}

\begin{table*}[pos=ht]
\centering
\small
\caption{Fixed configurations of the classical learning models used in the reference evaluation. Parameters not listed retain the \texttt{scikit-learn} 1.8.0 defaults. The same settings are used for the corresponding classifier and regressor unless stated otherwise.}
\label{tab:fixed_model_configurations}
\begin{tabular}{p{3.0cm}p{11.5cm}}
\toprule
\textbf{Model} & \textbf{Fixed configuration} \\
\midrule
Ridge &
Regularization strength $\alpha=1.0$; automatic solver selection; pseudorandom seed 42. \texttt{RidgeClassifier} is used for \ac{fc} and \texttt{Ridge} for \ac{fl}. \\

\ac{knn} &
Five neighbors; uniform weighting; Minkowski distance with $p=2$; automatic neighbor-search algorithm; parallel prediction enabled. \\

Histogram-based \ac{gb} &
100 boosting iterations; learning rate 0.1; unrestricted tree depth; minimum 20 samples per leaf; no L2 regularization; pseudorandom seed 42. \\

\ac{mlp} &
One hidden layer with 100 rectified-linear units; Adam optimizer; L2 penalty $10^{-4}$; initial learning rate $10^{-3}$; automatic batch size; maximum 200 iterations; no early stopping; pseudorandom seed 42. \\
\bottomrule
\end{tabular}
\end{table*}

This appendix reports the single-parameter hyperparameter ablations underlying the robustness summary in the main text. The benchmark results reported in the main analysis use the tagged default configurations; the ablations are diagnostic sensitivity checks and are not used as a hyperparameter-selection procedure for the reported models. In each ablation, one hyperparameter group is varied while all remaining settings are held fixed at the tagged default baseline. Table~\ref{tab:ablation_design_combined} defines the evaluated ablation design, and Table~\ref{tab:ablation_robustness_condensed} summarizes the resulting task- and window-specific sensitivity. The compact robustness analyses reported here are intended as diagnostic sensitivity checks; broader robustness stress testing of ML-based protection models is treated separately in~\cite{oelhaf_robustness_2026}.

\begin{table*}[pos=ht]
\centering
\small
\caption{Single-parameter hyperparameter ablation design.}
\label{tab:ablation_design_combined}
\begin{tabular}{@{}lll p{0.22\linewidth}@{}}
\toprule
\textbf{Model} & \textbf{Hyperparameter} & \textbf{Default} & \textbf{Tested values} \\
\midrule
\ac{gb} & Learning rate & 0.1 & 0.03, 0.2 \\
\ac{gb} & Maximum tree depth & unlimited & 3, 5, 10 \\
\ac{gb} & Boosting iterations & 100 & 50, 300 \\
\ac{gb} & Minimum samples per leaf & 20 & 5, 50 \\
\ac{gb} & L2 regularization & 0 & $10^{-4}$, $10^{-2}$ \\
\midrule
\ac{mlp} & Hidden layers & (100) & (50), (100, 50), (256, 128) \\
\ac{mlp} & L2 penalty & $10^{-4}$ & $10^{-5}$, $10^{-3}$ \\
\ac{mlp} & Initial learning rate & $10^{-3}$ & $10^{-5}$, $10^{-4}$, $10^{-2}$ \\
\ac{mlp} & Batch size & auto & 64, 128, 256 \\
\ac{mlp} & Training iterations & 200 & 100, 300, 400 \\
\bottomrule
\end{tabular}
\end{table*}

\begin{table*}[pos=ht]
\centering
\small
\caption{Condensed hyperparameter robustness summary across single-parameter ablations. For \ac{fc}, $\Delta_{\text{best}}$ denotes the increase in macro-\ac{f1}; for \ac{fl}, it denotes the reduction in \ac{mae}. Positive values therefore indicate improvement in both tasks. The maximum spread reports the largest best--worst difference within any single-parameter ablation group. The default cells are independent results from the ablation harness and can differ slightly from the Table~\ref{tab:reference_20ms} reference; for example, \ac{fl} \ac{mlp} at 20\,ms is $9.998$ here versus $10.20$ in the reference evaluation. Comparisons within each ablation group therefore use the corresponding default generated by the same harness. This difference is not a repeated-seed uncertainty estimate.}
\label{tab:ablation_robustness_condensed}
\begin{tabular}{@{}llcccll@{}}
\toprule
\textbf{Task} & \textbf{Model} & \textbf{Window} & \textbf{Default} & \textbf{Best metric} & \textbf{$\Delta_{\text{best}}$} & \textbf{Max. spread} \\
\midrule
\ac{fc} & \ac{gb}  & 20\,ms & 0.745  & 0.967 & +0.222 (min leaf) & 0.327 (learn. rate) \\
\ac{fc} & \ac{gb}  & 50\,ms & 0.982  & 0.988 & +0.006 (iters)    & 0.116 (learn. rate) \\
\ac{fc} & \ac{mlp} & 20\,ms & 0.990  & 0.996 & +0.006 (init. lr) & 0.015 (init. lr) \\
\ac{fc} & \ac{mlp} & 50\,ms & 0.991  & 0.994 & +0.003 (init. lr) & 0.011 (init. lr) \\
\midrule
\ac{fl} & \ac{gb}  & 20\,ms & 14.782 & 13.232 & +1.550 (iters)   & 5.530 (depth) \\
\ac{fl} & \ac{gb}  & 50\,ms & 14.662 & 12.942 & +1.720 (iters)   & 5.671 (depth) \\
\ac{fl} & \ac{mlp} & 20\,ms & 9.998 &  9.322 & +0.675 (hidden)  & 6.209 (init. lr) \\
\ac{fl} & \ac{mlp} & 50\,ms &  9.955 &  8.296 & +1.659 (hidden)  & 3.694 (init. lr) \\
\bottomrule
\end{tabular}
\end{table*}

\begin{table*}[pos=ht]
\caption{Compact hyperparameter robustness summary for the representative \ac{mlp} and histogram-based \ac{gb} models based on the single-parameter ablations reported in this appendix. Ablation default refers to the best tagged default result within the independent ablation campaign across the reported decision horizons. Best denotes the strongest result observed in the corresponding ablation setting. For \ac{fc}, $\Delta_{\text{best}}$ denotes the increase in macro-\ac{f1}; for \ac{fl}, it denotes the reduction in \ac{mae}. Positive values therefore indicate improvement in both tasks.}
\small
\label{tab:ablation_robustness_summary_main}
\centering
\begin{tabular}{llrrrr}
\toprule
\textbf{Model} & \textbf{Task} & \textbf{Ablation default} & \textbf{Best} & \textbf{$\Delta_{\text{best}}$} & \textbf{Max. spread} \\
\midrule
\ac{mlp} & \ac{fc} & 0.991 & 0.996 & +0.005 & 0.015 \\
\ac{gb} & \ac{fc} & 0.982 & 0.988 & +0.006 & 0.327 \\
\midrule
\ac{mlp} & \ac{fl} & 9.955 & 8.296 & +1.659 & 6.209 \\
\ac{gb} & \ac{fl} & 14.662 & 12.942 & +1.720 & 5.671 \\
\bottomrule
\end{tabular}
\end{table*}

\end{document}